\documentclass{article}
\usepackage{ijcai22}

\usepackage{times}
\usepackage{soul}
\usepackage{url}
\usepackage[hidelinks]{hyperref}
\usepackage[utf8]{inputenc}
\usepackage[small]{caption}
\usepackage{graphicx}
\usepackage{amsmath}
\usepackage{amsthm}
\usepackage{booktabs}
\usepackage{enumitem}
\usepackage{titlesec}
\usepackage{setspace}
\usepackage{multirow}
\usepackage{subfigure}
\usepackage{color}
\usepackage[linesnumbered, ruled]{algorithm2e}
\usepackage{amsfonts}
\usepackage[symbol]{footmisc}

\theoremstyle{definition}

\newcommand{\xhdr}[1]{{\noindent\bfseries #1}.}
\newcommand{\name}{GraphSANN\ }

\newcommand{\jie}[1]{{{\textcolor{black}{#1}}}}

\title{Imbalanced Node Classification Beyond Homophilic Assumption}

\author{
Jie Liu$^1$
\and
Mengting He$^1$\and
Guangtao Wang$^2$\and
Nguyen Quoc Viet Hung$^3$\and
Xuequn Shang$^{1\dagger}$\And
Hongzhi Yin$^{4\dagger}$
\affiliations
$^1$Northwestern Polytechnical University\\
$^2$Bytedance Inc\\
$^3$Griffith University\\
$^4$The University of Queensland
\emails
\{jayliu,hmt468\}@mail.nwpu.edu.cn,
xjtuwgt@gmail.com,
henry.nguyen@griffith.edu.au,
shang@nwpu.edu.cn,
h.yin1@uq.edu.au
}

\begin{document}

\maketitle
\begin{abstract}
    Imbalanced node classification widely exists in real-world networks where graph neural networks (GNNs) are usually highly inclined to majority classes and suffer from severe performance degradation on classifying minority class nodes. Various imbalanced node classification methods have been proposed recently which construct synthetic nodes and edges w.r.t. minority classes to balance the label and topology distribution. However, they are all based on the homophilic assumption that nodes of the same label tend to connect despite the wide existence of heterophilic edges in real-world graphs. Thus, they uniformly aggregate features from both homophilic and heterophilic neighbors and rely on feature similarity to generate synthetic edges, which cannot be applied to imbalanced graphs in high heterophily. To address this problem, we propose a novel \name for imbalanced node classification on both homophilic and heterophilic graphs. Firstly, we propose a \textit{unified feature mixer} to generate synthetic nodes with both homophilic and heterophilic interpolation in a unified way. Next, by randomly sampling edges between synthetic nodes and existing nodes as candidate edges, we design an \textit{adaptive subgraph extractor} to adaptively extract the contextual subgraphs of candidate edges with flexible ranges. Finally, we develop a \textit{multi-filter subgraph encoder} which constructs different filter channels to discriminatively aggregate neighbors' information along the homophilic and heterophilic edges. Extensive experiments on eight datasets demonstrate the superiority of our model for imbalanced node classification on both homophilic and heterophilic graphs.
\end{abstract}
\footnotetext[2]{Corresponding authors.}
\vspace{-3mm}
\section{Introduction}
Graph Neural Networks (GNNs) successfully extend deep learning approaches to graph data and have exhibited powerful learning ability on node classification task~\cite{kipf2016semi,velivckovic2017graph,liu2023meta}. Despite their effectiveness, most existing GNNs neglect the widely existing class-imbalance problem in real-world networks, where certain class(es) have significantly fewer node samples for training than other classes~\cite{sun2021multi}. For example, in online transaction networks, the majority of nodes are normal customers while only a small number are fraudsters; in molecular networks, there are much more low-mass atoms than high-mass atoms. Due to the dominating role of majority class nodes in the training set, classical GNNs are often highly inclined to majority classes, leading to severe performance degradation for minority node classification.

\begin{figure}[t]
    \centering
    \subfigure{
    \includegraphics[scale=0.5]{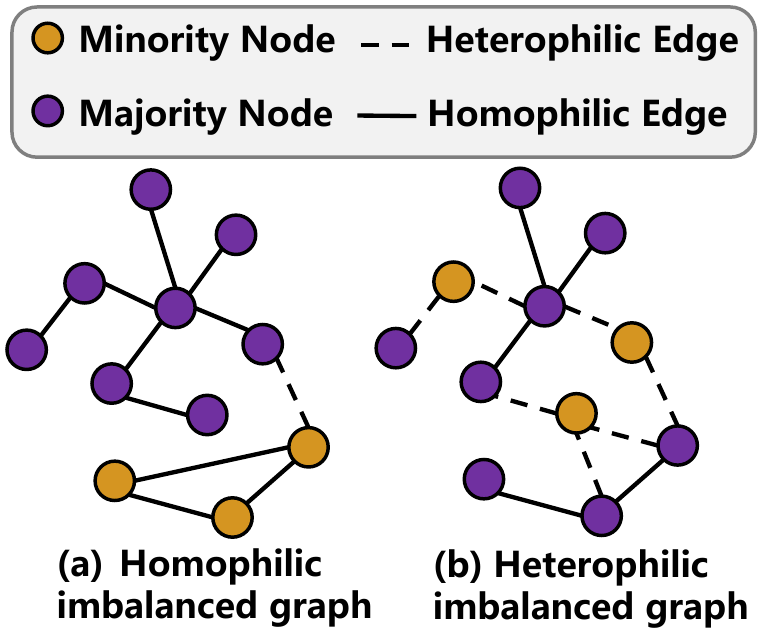}
    }\hspace{-1mm}
    \subfigure{
    \includegraphics[scale=0.33]{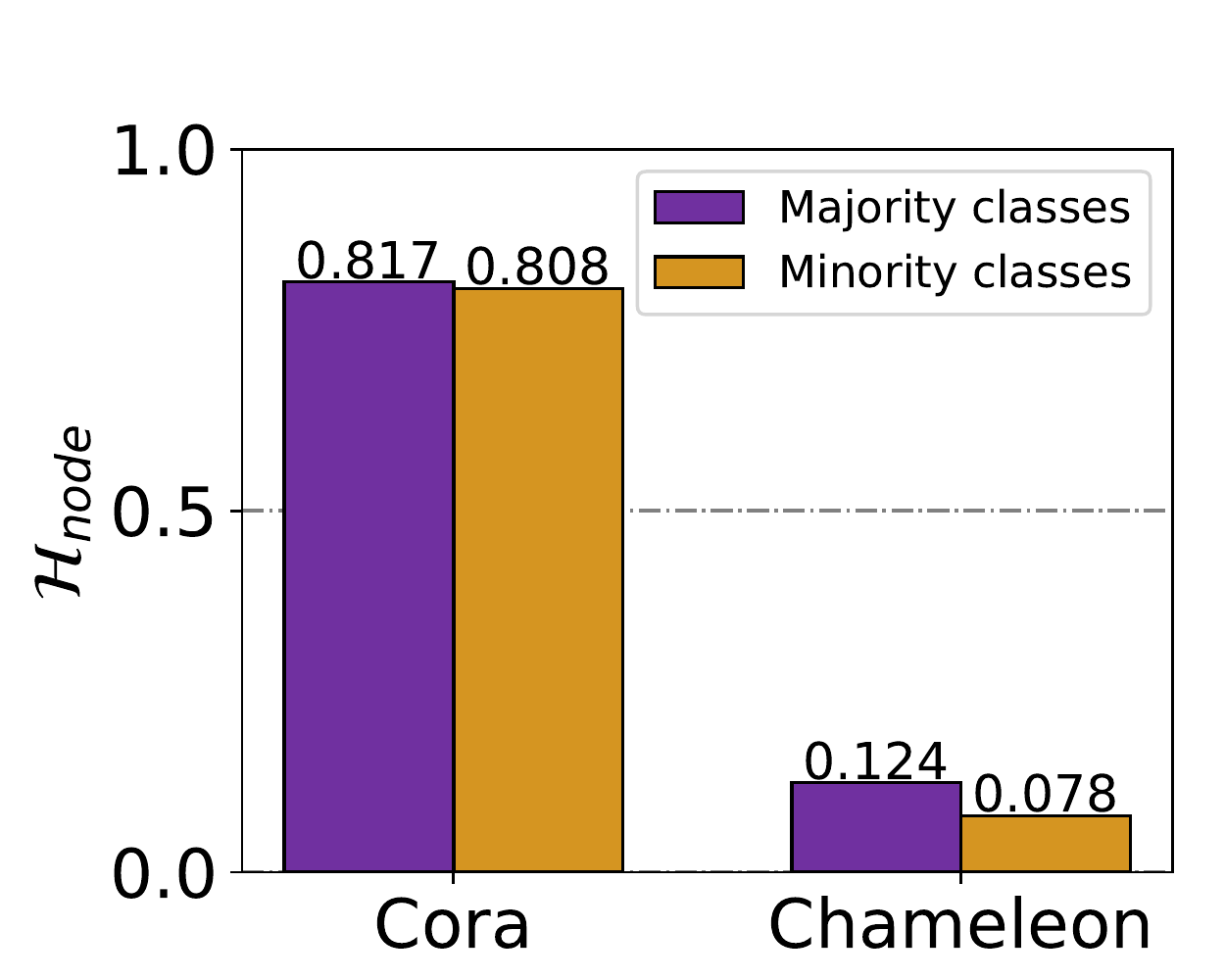}
    }
    \caption{\textbf{Left}: Illustration of (a) homophilic and (b) heterophilic imbalanced graphs. Many imbalanced networks exhibit strong heterophily. For example, in transaction networks, fraudsters often disguise themselves by connecting to normal customers. 
    \textbf{Right}: Comparison of node homophily $\mathcal{H}_{node}$ of a homophilic imbalanced network (Cora) and a heterophilic imbalanced network (Chameleon).}
    \label{fig:homo}
\end{figure}

To address the class-imbalance problem in the node classification task, many methods have been proposed recently.
\jie{They mainly relieve the imbalance problem by generating synthetic nodes for minority classes and further constructing synthetic edges between the generated nodes and the original nodes.} For example, GraphSMOTE~\cite{zhao2021graphsmote} generates synthetic nodes by interpolating nodes of the same minority class through SMOTE~\cite{chawla2002smote} and generates their linkages through a pre-trained edge generator. ImGAGN~\cite{qu2021imgagn} also synthesizes minority class nodes and connects them to real minority class nodes through a generative adversarial network. GraphENS~\cite{park2021graphens} further synthesizes the whole ego networks for synthetic minority nodes based on information from both minority and majority classes.

Although having acquired prominent performances on certain imbalanced datasets, these existing methods are based on the homophily assumption that edges tend to connect nodes of the same class label (Figure \ref{fig:homo}(a)). However, many investigations~\cite{sun2021heterogeneous,yu2020enhancing} show that heterophilic connections which link nodes of different classes also widely exist in imbalanced graphs (Figure \ref{fig:homo}(b)). Existing imbalanced node classification methods suffer from three severe problems when applied to networks with a large portion of heterophilic connections. \textbf{P1}: Most existing methods generate synthetic nodes based on homophilic interpolation, which restricts interpolated node pairs to be the same minority class. This causes synthetic nodes to lack diversity when real minority class nodes are very limited. \textbf{P2}: 
Existing models mainly resort to node feature similarity for synthetic edge construction. This strategy works well for homophilic edges which connect nodes with similar features but fail in constructing heterophilic edges and would thus introduce structure bias (i.e., heterophilic/homophilic edge distribution drift). \textbf{P3}: Existing methods conduct uniform message passing for both homophilic and heterophilic edges when aggregating features, and consequently result in much noisy information from dissimilar neighbors derived from heterophilic edges to be aggregated into the target nodes. This would seriously degrade the quality of node embeddings and hurt the following node classification task.

In light of this, we propose a novel \textbf{S}ubgraph-aware \textbf{A}daptive \textbf{Graph} \textbf{N}eural \textbf{N}etwork (\textbf{GraphSANN}) for imbalanced node classification on both homophilic and heterophilic graphs. \name consists of three major components, i.e., unified feature mixer, adaptive subgraph extractor, and multi-filter subgraph encoder. Specifically, to tackle \textbf{P1}, \name first applies a unified feature mixer to carry out both homophilic and heterophilic interpolation in a unified way. Next, to tackle \textbf{P2}, instead of generating edges based on feature similarity, we propose an adaptive subgraph extractor to extract the surrounding subgraphs of candidate synthetic edges with flexible ranges. In this way, distant but similar nodes can be absorbed into the subgraph whose general structural information will be encoded to predict the existence of the edge.
To tackle \textbf{P3} and encode the subgraphs consisting of both homophilic and heterophilic connections, we design a multi-filter subgraph encoder to aggregate messages only from similar nodes instead of dissimilar ones by fusing the output messages of three distinct filters. Finally, after generating synthetic nodes/edges and attaching them to the original graph, we apply a multi-filter GNN as node classifier to encode the acquired balanced graph for node classification.

The major contributions of this work are stated as follows:
\begin{itemize}[leftmargin=*]
    \item To the best of our knowledge, this paper is the first work to tackle the imbalanced node classification problem beyond the homophilic assumption.
    \item We design a novel imbalanced node classification model \name which is able to build balanced graph by generating synthetic nodes and constructing both homophilic and heterophilic synthetic edges between generated and original nodes, and aggregates the information from homophilic and heterophilic neighbors discriminatively.
    \item Extensive experiments on eight benchmark datasets show that \name acquires superior performance on both imbalanced homophilic and heterophilic graphs.
\end{itemize}

\vspace{-0.4cm}
\section{Related Work}
\xhdr{Heterophilic Graph Neural Networks}
Since most existing GNNs follow the homophily assumption and thus face significant performance degradation on heterophilic graphs, heterophily-based GNNs have been proposed, which can be roughly categorized into two groups~\cite{zheng2022graph}: (1) Neighbor extension methods which aim to expand local neighborhood to absorb features from distant but informative nodes. For example, MixHop~\cite{abu2019mixhop} aggregates messages from multi-hop neighbors respectively and mixes them together through concatenation. UGCN~\cite{jin2021universal} further restricts nodes from two-hop neighbors to have at least two different paths to the ego node. (2) Adaptive message aggregation methods which design adaptive aggregation operations to learn discriminative information from homophilic and heterophilic linkages. For example, FAGCN~\cite{bo2021beyond} adopts a self-gating attention mechanism to uniformly learn low-frequency and high-frequency signals from neighbors. ACM~\cite{luan2022revisiting} further designs a linear combination of low-pass and high-pass filters to adaptively learn information from different filter channels. 

\xhdr{Imbalanced Node Classification}
Generally, imbalanced node classification methods can be divided into two groups, generic and network-specific methods. Generic ones directly combine general class-imbalance approaches (e.g. oversampling, re-weight, SMOTE~\cite{chawla2002smote}, etc.) with GNNs to graph data. For example, \textit{Oversampling}~\cite{buda2018systematic} replicates existing node embeddings learned from GNNs to produce more minority node representations; \textit{Re-weight}~\cite{yuan2012sampling+} assigns larger penalty weights to minority nodes when computing training loss. 
Network-specific methods usually take account of the sophisticated topology of graphs to generate synthetic nodes and further determine the connections between the generated nodes and original nodes~\cite{chen2021topology,xia2021self}.
Among them, DR-GCN~\cite{shi2020multi} utilizes class-conditional adversarial training to enhance the separation of different classes. GraphSMOTE~\cite{zhao2021graphsmote} interpolates nodes from minority classes as synthetic nodes and generates linkages through a pre-trained edge generator. ImGAGN~\cite{qu2021imgagn} synthesizes minority nodes and connects them to original nodes through a generative adversarial network. GraphENS~\cite{park2021graphens} demonstrates the overfitting problem of neighbor memorization and proposes to generate the whole ego-networks for synthetic nodes based on information from all classes.

However, these models are all based on the homophily assumption and thus suffer from performance degradation when applied to networks with strong heterophily.

\vspace{-0.2cm}
\section{Notations and Problem Definition}
\vspace{-0.1cm}
\xhdr{Definition 1. Graph Homophily and Heterophily}
Given a graph $\mathcal{G}=\{\mathcal{V}, \mathcal{E}\}$, where $\mathcal{V}$ represents the node set, and for each edge $e=(v, \mu), v, \mu \in \mathcal{V}$, if $v$ and $\mu$ have the same class label, the edge $e$ is homophilic. Otherwise, $e$ is heterophilic. Most graphs have both homophilic and heterophilic edges at the same time. We define the node homophily  $\mathcal{H}_{node}$ and edge homophily $\mathcal{H}_{edge}$ to quantitatively measure the homophily degree of a graph as follows. 
\vspace{-0.2cm}
\begin{subequations}
  \begin{flalign}
    \mathcal{H}_{node} &= \frac{1}{|\mathcal{V}|}\sum_{v\in\mathcal{V}}\frac{|\mu \in \mathcal{N}(v):y_v=y_{\mu}|}{|\mathcal{N}(v)|}, \\
    \mathcal{H}_{edge} &= \frac{|\{(v,\mu)\in \mathcal{E}:y_v=y_\mu\}|}{|\mathcal{E}|}.
    \end{flalign}
\end{subequations}
Based on the definitions of $\mathcal{H}_{node}, \mathcal{H}_{edge} \in [0, 1]$, the node heterophily and edge heterophily can be defined as $1 -\mathcal{H}_{node}$ and $1-\mathcal{H}_{edge}$, respectively. Graphs with strong homophily have higher $\mathcal{H}_{node}$ and $ \mathcal{H}_{edge}$, and vice versa.


\xhdr{Problem Definition}
Let $\mathcal{G}=\{\mathcal{V}, \mathcal{E}, \mathbf{X}\}$ denote an attribute graph, where $\mathbf{X}\in \mathbb{R}^{|\mathcal{V}|\times d}$ is node feature matrix whose $i$-th row represents a $d$-dimensional feature vector of the $i$-th node. For node classification, each node is also associated with a one-hot node label $Y_{i,:}\in \mathbb{R}^{C}$ where $C$ is the number of node classes. 
If the class size distribution is imbalanced, we name it as an imbalanced node classification problem. The imbalance ratio is defined as \textit{im\_ratio} = $\frac{min_c(|\mathcal{V}_c|)}{max_c(|\mathcal{V}_c|)}\ll 1$, where $\mathcal{V}_c|$ denotes the node set with class label $c \in \{1, 2, \cdots, C\}$. Then, we give the formal definition of imbalanced node classification as follows.

 \textit{Given an imbalanced graph $\mathcal{G}=\{\mathcal{V}, \mathcal{E}, \mathbf{X}\}$ composed of both homophilic and heterophilic edges, our goal is to learn a node classifier $f:f(\mathcal{V},\mathcal{E},\mathbf{X})\to \mathbf{Y}$ that can well classify both majority and minority classes and can be well generalized on graphs with either low or high heterophily.}
\vspace{-2mm}
\section{Methodology}
\begin{figure*}[t]
  \centering
  \includegraphics[scale=0.72]{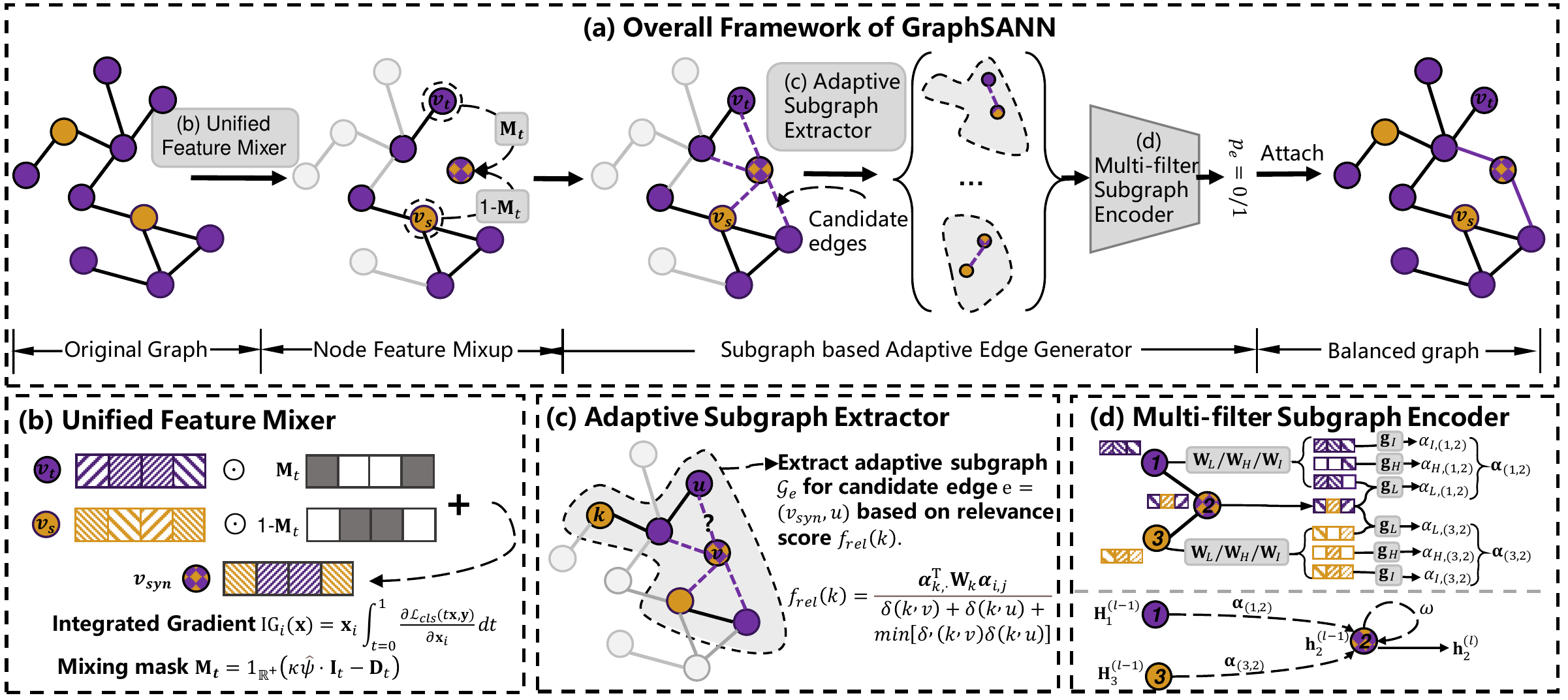}
  \caption{(a) Overall framework of \name. It is composed of three core components, i.e., (b) unified feature mixer (c) adaptive subgraph extractor, and (d) multi-filter subgraph encoder.} \label{fig:framework}
\end{figure*}

In this section, we introduce our novel \name for imbalanced node classification. As illustrated in Figure \ref{fig:framework}, \name consists of three core components, (1) \textit{Unified Feature Mixer} which carries out both homophilic and heterophilic interpolation to generate synthetic nodes (Subsection \ref{sub:mixer}); (2) \textit{Adaptive Subgraph Extractor} which adaptively extracts subgraphs around candidate synthetic edges (Subsection \ref{sub:extractor}) and (3) \textit{Multi-filter based subgraph encoder} which encodes the subgraphs extracted from component (2) into edge score using multiple passes of filters (Subsection \ref{sub:encoder}). Please refer to Algorithm \ref{alg:1} in the Appendix for the forward propagation procedure of \name. We elaborate on each component in the following subsections. 

\vspace{-1mm}
\subsection{Unified Feature Mixer} \label{sub:mixer}
Most existing class-imbalance models apply homophilic interpolation on existing node pairs $<v_s,v_t>$ to generate synthetic nodes, which restricts $v_s$ and $v_t$ to be the nearest neighbors of the same minority class~\cite{zhao2021graphsmote,qu2021imgagn}. The generated nodes under this strategy suffer from feature diversity problem especially when original minority nodes are very limited~\cite{park2021graphens}. To address this problem, we propose a unified feature mixer that conducts both homophilic and heterophilic interpolation in a unified way, which consists of two steps: (1) unified node pair sampling and (2) integrated gradient based feature mixup.

\xhdr{Unified Node Pair Sampling}
Before conducting feature mixup to generate synthetic minority nodes, we first sample node pairs $<v_s, v_t>$ from the existing node set for interpolation. Here $v_s$ is sampled from minority classes while $v_t$ is from the entire classes, and thus it could have either the same or different class compared to $v_s$, formed as $S_{pair}=\{<v_s, v_t>\mid v_s\in\mathcal{V}_{minor},\, v_t\in\mathcal{V}\}$. Let $\mathcal{C}=\{1,2,\ldots, C\}$ and $\mathcal{C}_M \subset \mathcal{C}$ be the entire class set and minority class set, the sampling distributions for $v_s$ and $v_t$ are denoted as $v_s \sim p_s(u\mid \mathcal{C}_M)$ and $v_t \sim p_t(u\mid \mathcal{C})$, respectively. To obtain a uniformly sampling  $v_t$ from all the classes, we define $p_s$ and $p_t$ as follows:
\vspace{-3mm}
\begin{subequations}
  \begin{flalign}
    p_s(u\mid \mathcal{C}_M) &= \frac{1}{|\mathcal{V}_{m}|}, \  m \in \mathcal{C}_M, \\
    p_t(u\mid \mathcal{C}) &= \frac{\text{log}(|\mathcal{V}_c| + 1)}{(|\mathcal{V}_c|+1)\sum_{c\in\mathcal{C}}\text{log}(|\mathcal{V}_c|+1)}.
    \end{flalign}
\end{subequations}
It is straightforward to get that $p_t$ reaches the peak when $|\mathcal{V}_c|= 2$ and gets 0 when $|\mathcal{V}_c|=0$ or $|\mathcal{V}_c|=\infty$. In this way, even if the sizes of majority classes are significantly larger than minority classes, nodes from the entire classes have roughly equal chances to be sampled as $v_t$. Here we use over-sampling scale $\zeta$ to control the amount of sampled node pairs.

\xhdr{Integrated Gradient based Feature Mixup}
Afterward, we interpolate the raw features of each node pair $<v_s, v_t>$ to generate synthetic nodes. Since $v_t$ is sampled beyond the same minority class as $v_s$, we only preserve generic node attributes which are irrelevant to class prediction to avoid introducing distracting information. As introduced by~\cite{sundararajan2017axiomatic}, integrating gradient can effectively evaluate the contributions of input features to the model prediction. Compared to directly using gradients to evaluate feature importance~\cite{park2021graphens}, integrated gradient addresses the saturation and thresholding problems~\cite{shrikumar2017learning} and can acquire more reliable feature importance. Specifically, the integrated gradient $\text{IG}_i(\mathbf{x})$ along the $i$-th dimension of input node feature $\mathbf{x}\in\mathbb{R}^d$ is calculated as follows:
\begin{equation}\label{eq:integral}
    \text{IG}_i(\mathbf{x}) = \mathbf{x}_i\int^1_{t=0}\frac{\partial \mathcal{L}_{cls}(t\mathbf{x},\mathbf{y})}{\partial\mathbf{x}_i}dt,
\end{equation}
where $\mathbf{y}$ represents the vector of true class labels and $\mathcal{L}_{cls}$ denotes the node classification loss. Then, we compute the distance $\psi_{st}$ between $v_s$ and $v_t$ by:
\begin{equation}
    \psi_{st} = \large\Vert \mathbf{W}_p\mathbf{x}_s-\mathbf{W}_p\mathbf{x}_t \large\Vert_{2},
\end{equation}
where $\mathbf{W}_p\in\mathbb{R}^{d\times d'}$represents the projection matrix, and $\mathbf{x}_s$ and $\mathbf{x}_t$ are raw attributes of $v_s$ and $v_t$, respectively. We define $\hat{\psi}_{st} = \frac{1}{1+\psi_{st}} \in [0,1]$ as the similarity between $v_s$ and $v_t$. Finally, we construct the masking vector $\mathbf{M}_t\in\mathbb{R}^d$ as follows:
\begin{equation}\label{eq:mv}
    \mathbf{M}_t = 1_{\mathbb{R}^+}(\kappa\hat{\psi_{st}}\cdot\mathbf{I}_t-\mathbf{D}_t)),
\end{equation}
where $\mathbf{I}_t \in \mathbb{R}^{d}$ is an all-ones vector, $\kappa$ is a hyper-parameter, $\mathbf{D}_t=[\text{IG}_1(\mathbf{x}_t),\ldots,\text{IG}_d(\mathbf{x}_t)]$ is the feature importance vector for $x_t$, and $1_{\mathbb{R}^+}(\cdot)$ is an indicator function which returns 1 when the input is positive otherwise 0. Thus, the mixed feature $\mathbf{x}_{syn}$ of synthetic nodes is formulated as follows:
\begin{equation}\label{eq:x_syn}
    \mathbf{x}_{syn} = (1-\mathbf{M}_t)\odot \mathbf{x}_s + \mathbf{M}_t\odot\mathbf{x}_t.
\end{equation}

\vspace{-2mm}
\subsection{Adaptive Subgraph Extractor} \label{sub:extractor}
After generating synthetic nodes $v_{syn}\in\mathcal{V}_{syn}$ for the minority classes, we need to determine their connections to the original graph. The existence of an edge $e=(v_{syn},u)$ is highly related to the structural information embedded in its surrounding subgraph $\mathcal{G}_{e}=\{\mathcal{V}_e,\mathcal{E}_e\}$ regardless of its homophily or heterophily~\cite{zhang2018link}. 
Here, we first randomly sample connections between synthetic and original nodes as candidate edges, then we adaptively extract the subgraphs of candidate edges and finally encode their structure information to predict edge existence.
Specifically, given the node pair set $S_{pair}$ and synthetic node set $\mathcal{V}_{syn}$, the candidate synthetic edge set $\mathcal{E}_{syn}$ is constructed as follows:
\begin{align}
    \mathcal{E}_{syn} &= \{(v,u)\mid v\in\mathcal{V}_{syn}, u\in\mathcal{V}_{nei}\}, \\
    \mathcal{V}_{nei} &=\left[\mathcal{N}_{1}(v_s)\cup \mathcal{N}_{1}(v_t)\right]_{\xi}, <v_s, v_t>\in S_{pair},
\end{align}
where $\mathcal{N}_{1}(v)$ returns 1-hop neighbors of $v$ and $v$ itself, $[\cdot]_{\xi}$ is random sampling with sampling ratio $\xi$. After obtaining $\mathcal{E}_{syn}$, we extract the enclosing subgraph $\mathcal{G}_e$ for each $e\in\mathcal{E}_{syn}$. Considering that nodes with high structural and semantic similarities might be distant from each other in heterophilic graphs, instead of fixing the subgraph to $h$-hop neighbors, we adaptively adjust the range of a subgraph based on a relevance score function $f_{rel}(\cdot)$. For any $h$-hop neighbor $ k \in \mathcal{N}_h(v)\cup\mathcal{N}_h(u)$, $f_{rel}(k)$ is calculated as follows:
\begin{align}
     f_{rel}(k) &= \frac{\boldsymbol{\alpha}_{k,\cdot}^{(l),\mathrm{T}}\mathbf{W}_k\boldsymbol{\alpha}^{(l)}_{i,j}}{\delta(k,v)+ \delta(k,u)+min[\delta(k,v), \delta(k,u)]}, \\
     \boldsymbol{\alpha}^{(l)}_{k,\cdot} &= \frac{1}{|\mathcal{N}_1(k)|}\sum_{i\in\mathcal{N}_1(k)}\boldsymbol{\alpha}^{(l)}_{k,i}.
\end{align}
Where $\delta(k,v)$ is the length of shortest path from $k$ to $v$, $\boldsymbol{\alpha}_{k,i}\in \mathbb{R}^3$ represents the multi-pass weight vector between node $k$ and $i$ (further illustrated in Subsection \ref{sub:encoder}). $\boldsymbol{\alpha}^{(l)}_{k,\cdot}$ is the mean value of all the coefficient vectors between $k$ and its 1-hop neighbors, which reflects the homophily status of node $k$. $\mathbf{W}_k\in\mathbb{R}^{3\times3}$ is a weight matrix. 
Then we select top $M$ nodes from $f_{rel}(k)$ along with central nodes $u$ and $v$ to construct the enclosing subgraph. The $M$ is calculated based on subgraph density: $M = \lceil|\mathcal{V}_{e}|(1+\frac{2|\mathcal{E}_e|}{|\mathcal{V}_e|(|\mathcal{V}_e-1|)})\rceil$.

\vspace{-2mm}
\subsection{Multi-filter Subgraph Encoder}\label{sub:encoder}
\vspace{-1mm}
In this section, we propose a novel subgraph encoder to embed the subgraph surrounding a candidate edge $e \in \mathcal{E}_{syn}$ into a vector, and then predict whether the edge $e$ should be generated or not based on that. Let $\mathcal{G}_{e}$ be the extracted enclosing subgraph of the candidate edge $e$. Considering the widely existing heterophilic connections in $\mathcal{G}_e$, we design a multi-filter subgraph encoder that can discriminatively aggregate information from homophilic and heterophilic neighbors. Specifically, let $\mathbf{h}^{(l-1)}_u\in\mathbb{R}^{d_{l-1}\times1}$ denote the $(l-1)$-th layer feature of node $u \in \mathcal{G}_e, \mathbf{h}^{(0)}_u=[\mathbf{x}_u\parallel\mathbf{z}_u]$ where $\mathbf{z}_u$ is one-hot labeling feature acquired by DRNL\cite{zhang2018link}. We now compute the weight coefficients $\alpha^{(l)}_{L,(u,k)}, \alpha^{(l)}_{H,(u,k)}, \alpha^{(l)}_{I,(u,k)}$ which reflect the importance of different frequencies of signals as follows:
\begin{align}
     \alpha^{(l)}_{L,(u,k)} &= \sigma\Big(\mathbf{g}^\mathrm{T}_L\big[\mathbf{W}^{(l)}_L\mathbf{h}^{(l-1)}_u\parallel\mathbf{W}^{(l)}_L\mathbf{h}^{(l-1)}_k\big]\Big),\\
     \alpha^{(l)}_{H,(u,k)} &= \sigma\Big(\mathbf{g}^\mathrm{T}_H\big[-\mathbf{W}^{(l)}_H\mathbf{h}^{(l-1)}_k\big]\Big), \\
     \alpha^{(l)}_{I,(u,k)} &= \sigma\Big(\mathbf{g}^\mathrm{T}_I\big[\mathbf{W}^{(l)}_I\mathbf{h}^{(l-1)}_u\big]\Big).
\end{align}
Where $\mathbf{W}^{(l)}_L$,$\mathbf{W}^{(l)}_{H}$, $\mathbf{W}^{(l)}_{I}\in\mathbb{R}^{d_{l}\times d_{l-1}}$ are the weight matrices that project $\mathbf{h}^{(l-1)}_u$ into low-frequency, high-frequency and identity messages, respectively. $\mathbf{g}^\mathrm{T}_L\in\mathbb{R}^{2d_l\times1}$, $\mathbf{g}^\mathrm{T}_H$, $\mathbf{g}^\mathrm{T}_I\in \mathbb{R}^{d_l\times1}$ are the convolutional vectors, $\sigma(\cdot)$ is the Sigmoid function. Then we compute weight vector $\boldsymbol{\alpha}^{(l)}_{(u,k)}$ by normalizing the importance weights of different frequencies:
\begin{align}
    \boldsymbol{\alpha}^{(l)}_{(u,k)} &= \big[\widetilde{\alpha}^{(l)}_{L,(u,k)}, \widetilde{\alpha}^{(l)}_{H,(u,k)}, \widetilde{\alpha}^{(l)}_{I,(u,k)}\big], \\
    \widetilde{\alpha}^{(l)}_{i,(u,k)} &= \frac{\text{exp}\big(\alpha^{(l)}_{i,(u,k)}\big)}{\sum_{i\in\{L,H,I\}}\text{exp}\big(\alpha^{(l)}_{i,(u,k)}\big)}.
\end{align}
Next, we aggregate multi-frequency messages of neighbor nodes $k$ with weight vector $\boldsymbol{\alpha}^{(l)}_{(u,k)}$ to compute the central node embedding $\mathbf{h}^{(l)}_u$:
\begin{align}\small
    \mathbf{h}^{(l)}_u &= \omega\mathbf{h}^{(l-1)}_u+\sum_{k\in \mathcal{N}_1(u)}\boldsymbol{\alpha}^{(l)}_{(u,k)}\mathbf{H}^{(l-1)}_k,\\
    \mathbf{H}^{(l-1)}_k &= \text{ReLU}\Big(\big[\mathbf{W}^{(l)}_L\mathbf{h}^{(l-1)}_k,\mathbf{W}^{(l)}_H\mathbf{h}^{(l-1)}_k,\mathbf{W}^{(l)}_I\mathbf{h}^{(l-1)}_k\big]^{\mathrm{T}}\Big).
\end{align}
Where $\omega$ is a hyper-parameter, $\mathbf{h}^{(L)}_u$ denotes the aggregated node embedding of node $u\in \mathcal{G}_e$ after stacking $L$-layer encoders. To acquire structural information from different orders of neighbors, we concatenate node embeddings from different layers and utilize a mean readout to compute the existence probability $p_e$ of subgraph $\mathcal{G}_e$:
\begin{align}
    \mathbf{h}_\mu &= \mathbf{h}_\mu^{(1)}\parallel\mathbf{h}_\mu^{(2)}\parallel\ldots\parallel\mathbf{h}_\mu^{(L)}, \\
    p_e &= \frac{1}{|\mathcal{V}_e|}\sum_{u\in\mathcal{V}_e}\mathbf{W}_{pool}\mathbf{h}_u.
\end{align}
Where $\mathbf{W}_{pool}\in \mathbb{R}^{1\times (d_1+\ldots+d_L)}$ projects the latent embeddings into a scalar $p_e$ which reflects the existence probability of synthetic edge $e$. We remove the edges with low $p_e$ from $\mathcal{E}_{syn}$ based on threshold $\eta$ and attach the rest of synthetic edges to the original graph to construct the adjacency matrix $\tilde{\mathbf{A}}$ after over-sampling:
  \begin{equation}
  \centering
    \tilde{\mathbf{A}}(v_{syn},u) =
    \begin{cases}
    1,  & \text{if $p_e > \eta$} \\
    0, & \text{otherwise.} \nonumber
    \end{cases}
  \end{equation}
  
\subsection{Optimization Objective}
In this section, we introduce the optimization objective of our proposed \name for imbalanced node classification, which consists of two optimization tasks: 1) adjacency matrix reconstruction and 2) node classification.

\xhdr{Adjacency Matrix Reconstruction} We train our multi-filter subgraph encoder with an adjacency matrix reconstruction task. Let $\mathbf{A}$ denote the adjacency matrix of the original graph, and $\mathbf{A}(u,v)=1$ indicate the existence of an edge between $u$ and $v$. Considering the sparsity of positive edges, we also adopt negative sampling~\cite{zhou2022link}. Specifically, for each positive edge $\mathbf{A}(u,v)=1$, we randomly sample an unlinked edge which makes $\mathbf{A}(u,m)=0$ as a negative sample and constructs a negative set $\mathcal{M}^{-}$. The loss function for adjacency matrix reconstruction is formed as follows:
\begin{equation}\small
    \mathcal{L}_{rec} =\sum_{\mathbf{A}(u,v)>0,
    \atop
    (v,m)\in\mathcal{M}^-}\Big[\big\Vert\hat{\mathbf{A}}(u,v)-\mathbf{A}(u,v)\big\Vert^2_F+\big\Vert\hat{\mathbf{A}}(u,m)-\mathbf{A}(u,m)\big\Vert^2_F\Big],
\end{equation}
where $\hat{\mathbf{A}}$ is the predicted adjacency matrix of original graph. 

\xhdr{Node classification} After attaching the synthetic nodes and edges to the original graph, we transform it into a balanced network. Since the balanced graph can also be heterophilic, we adopt the multi-filter graph encoder introduced in \ref{sub:encoder} as node classifier by replacing the readout procedure with a one-layer MLP followed by Softmax:
\begin{equation}
    \hat{\mathbf{y}}_u = \text{Softmax}(\text{MLP}(\mathbf{h}^{(L)}_{u})).
\end{equation}
The output dimension of $\text{MLP}(\cdot)$ is equal to class number $C$. The loss function of node classification is defined as follows:
\begin{equation}
    \mathcal{L}_{cls} = -\frac{1}{|\mathcal{V}|}\sum^{|\mathcal{V}|}_{v=1}\sum_{c=1}^{C}\text{log}(\hat{\mathbf{y}}_v[c]\cdot\mathbf{y}_v[c])
\end{equation}
The overall objective function is then formed as follows with $\lambda \in $ (0, 1]:
\begin{equation}
    \textbf{min} \mathcal{L}=(1-\lambda)\mathcal{L}_{rec} + \lambda\mathcal{L}_{cls}.
\end{equation}

\begin{table}[t]\small
  \setlength\tabcolsep{2.5pt} 
  \begin{tabular}{l|rrrrrrr}
    \toprule
    \textbf{Datasets} & \textbf{Nodes} & \textbf{Edges} & \textbf{Features} & \textbf{Classes}& $\boldsymbol{\mathcal{H}_{\textbf{edge}}}$ & $\boldsymbol{\mathcal{H}_{\textbf{node}}}$ \\
    \midrule
    Cora &2,708 &5,429 &1,433 &7 &0.8100 &0.8252  \\
    Pubmed &19,717 &44,338 &500 &3 &0.8024 &0.7924 \\
    Citeseer  &2,277 &4,732 &3,703 &6 &0.7362 &0.7175  \\
    Chameleon  &5,201 &36,101 & 2,325 & 5&0.2795&0.2470  \\
    Squirrel &5,201 &217,073 & 2,089&5 & 0.2416&0.2156  \\
    Film &7,600 &33,544 &931 &5 &0.2200 &0.2400  \\
    Amazon-CP &13,381 &245,778 &767 &10 &0.7721 & 0.7853 \\
    Amazon-PH &7,487 &119,043 &745 &8 &0.8272 &0.8365 \\
    \bottomrule
  \end{tabular}
  \caption{Statistics of Datasets.}
  \label{tab:datasets}
\end{table}

\begin{table*}[!h]\scriptsize
  \setlength\tabcolsep{2.0pt} 
  \centering 
  \begin{tabular}{l|ccc|ccc|ccc|ccc}
    \toprule
    \textbf{Method}&\multicolumn{3}{c|}{\textbf{Cora}} &\multicolumn{3}{c|}{\textbf{Pubmed}} & \multicolumn{3}{c|}{\textbf{Citeseer}} & \multicolumn{3}{c}{\textbf{Amazon-Computers}} \\ 
    \textbf{$\mathcal{H}_{edge}$}&\multicolumn{3}{c|}{0.8100} &\multicolumn{3}{c|}{0.8024} & \multicolumn{3}{c|}{0.7362} & \multicolumn{3}{c}{0.7721}\\ 
    \midrule
    Metrics($\%$)& ACC & F1 & AUC & ACC & F1 & AUC & ACC & F1 & AUC & ACC & F1 & AUC \\
    \midrule
    GCN &53.68\scriptsize{$\pm$1.61} &45.63\scriptsize{$\pm$0.35} &81.30\scriptsize{$\pm$0.62} &53.69\scriptsize{$\pm$0.49} &51.66\scriptsize{$\pm$2.53} &71.86\scriptsize{$\pm$0.45} &44.59\scriptsize{$\pm$0.60} &28.38\scriptsize{$\pm$1.37} &74.35\scriptsize{$\pm$0.63} &55.32\scriptsize{$\pm$0.35} &44.16\scriptsize{$\pm$0.36} & 89.80$\pm$1.79 \\
    ACM &55.28\scriptsize{$\pm$0.75} &47.95\scriptsize{$\pm$0.48} &85.23\scriptsize{$\pm$0.48} &49.72\scriptsize{$\pm$0.73} &50.43\scriptsize{$\pm$1.28} &68.25\scriptsize{$\pm$1.02} &48.32\scriptsize{$\pm$0.58} &30.47\scriptsize{$\pm$1.03} &78.56\scriptsize{$\pm$0.47} &56.32\scriptsize{$\pm$0.43} &46.13\scriptsize{$\pm$0.47} & 91.26$\pm$1.78 \\
    Oversampling &62.79\scriptsize{$\pm$0.79}&52.06\scriptsize{$\pm$0.51} &89.48\scriptsize{$\pm$0.78} &61.15\scriptsize{$\pm$0.37} &60.33\scriptsize{$\pm$0.36} &78.67\scriptsize{$\pm$1.38} &51.05\scriptsize{$\pm$0.43} &32.86\scriptsize{$\pm$1.26} &82.99\scriptsize{$\pm$0.69} &55.39\scriptsize{$\pm$0.31} &44.31\scriptsize{$\pm$1.07} &88.03 $\pm$ 1.59\\
    Re-weight &63.16\scriptsize{$\pm$1.53} &52.39\scriptsize{$\pm$0.32} &90.16\scriptsize{$\pm$0.51} &62.21\scriptsize{$\pm$0.44} &61.12\scriptsize{$\pm$0.54} &79.02\scriptsize{$\pm$2.41} &50.91\scriptsize{$\pm$0.35} &32.79\scriptsize{$\pm$0.65} &82.86\scriptsize{$\pm$1.38} &56.78\scriptsize{$\pm$0.69} &48.12\scriptsize{$\pm$1.25} & 91.12 $\pm$ 1.84\\
    DR-GCN & 67.77\scriptsize{$\pm$1.09} & 67.67\scriptsize{$\pm$0.74} & 87.23\scriptsize{$\pm$0.28} & 55.33\scriptsize{$\pm$0.23} & 46.56\scriptsize{$\pm$0.43} & 67.45\scriptsize{$\pm$1.01} & 46.84\scriptsize{$\pm$1.42} & 34.54\scriptsize{$\pm$1.33} & 72.48\scriptsize{$\pm$0.83} &24.86\scriptsize{$\pm$1.27} &30.93\scriptsize{$\pm$1.76} &64.53 $\pm$ 2.11  \\
    ImGAGN & 63.60\scriptsize{$\pm$0.55} & 62.89\scriptsize{$\pm$0.60} & 91.87\scriptsize{$\pm$0.53} & 63.21\scriptsize{$\pm$1.25} & 62.13\scriptsize{$\pm$0.87} &78.32\scriptsize{$\pm$2.34} & 48.04\scriptsize{$\pm$0.78} & 36.14\scriptsize{$\pm$1.01} & 80.61\scriptsize{$\pm$0.21} &60.69\scriptsize{$\pm$1.25} &42.55\scriptsize{$\pm$1.91} & 91.25 $\pm$ 0.39 \\
    GraphSMOTE &66.76\scriptsize{$\pm$0.80} &65.86\scriptsize{$\pm$0.81} &93.75\scriptsize{$\pm$0.23} &64.98\scriptsize{$\pm$1.70} &64.05\scriptsize{$\pm$2.12} &81.62\scriptsize{$\pm$2.75} &48.20\scriptsize{$\pm$0.81} &34.65\scriptsize{$\pm$0.51} &77.72\scriptsize{$\pm$0.43} & 70.02\scriptsize{$\pm$0.98}&62.01\scriptsize{$\pm$0.85} & 96.26 $\pm$ 0.04\\
    GraphENS &72.68 \scriptsize{$\pm$0.76}& 67.94\scriptsize{$\pm$0.94} &94.32\scriptsize{$\pm$0.54} & 69.98 \scriptsize{$\pm$2.41} & 69.53 \scriptsize{$\pm$2.31} & 87.46\scriptsize{$\pm$ 1.58}& 53.18 \scriptsize{$\pm$2.90} & 49.48 \scriptsize{$\pm$3.28} &83.52\scriptsize{$\pm$2.14} &83.20\scriptsize{$\pm$0.27} &80.59\scriptsize{$\pm$0.37} &98.13 $\pm$ 0.06 \\
    \midrule
    \textbf{\name}&\textbf{77.73}\scriptsize{$\pm$0.75} & \textbf{74.94}\scriptsize{$\pm$0.29} & \textbf{95.59}\scriptsize{$\pm$0.33} & \textbf{75.54}{$\pm$1.12} & \textbf{74.81}\scriptsize{$\pm$0.65} & \textbf{90.68}\scriptsize{$\pm$0.40} &\textbf{66.39} \scriptsize{$\pm$0.15} &\textbf{61.97} \scriptsize{$\pm$0.14} &\textbf{85.62} \scriptsize{$\pm$0.72}  &\textbf{85.68}\scriptsize{$\pm$0.43} &\textbf{84.21}\scriptsize{$\pm$0.65} &\textbf{99.69}\scriptsize{$\pm$0.33} \\ 
    \bottomrule
    \toprule
    \textbf{Dataset}&\multicolumn{3}{c|}{\textbf{Chameleon}} & \multicolumn{3}{c|}{\textbf{Film}} & \multicolumn{3}{c|}{\textbf{Squirrel}} & \multicolumn{3}{c}{\textbf{Amazon-Photo}} \\
    \textbf{$\mathcal{H}_{edge}$} &\multicolumn{3}{c|}{0.2795} & \multicolumn{3}{c|}{0.2516} & \multicolumn{3}{c|}{0.2416} & \multicolumn{3}{c}{0.8272} \\
    \midrule
    Metrics($\%$)& ACC & F1 & AUC & ACC & F1 & AUC & ACC & F1 & AUC & ACC & F1 & AUC \\
    \midrule
    GCN &36.40\scriptsize{$\pm$2.14} &26.47\scriptsize{$\pm$1.91} &61.75\scriptsize{$\pm$2.33} & 23.39\scriptsize{$\pm$1.12} &17.09 \scriptsize{$\pm$1.33} & 55.39\scriptsize{$\pm$1.07} &22.69\scriptsize{$\pm$1.06} &17.43\scriptsize{$\pm$0.87} &49.51\scriptsize{$\pm$1.11} &67.23\scriptsize{$\pm$1.98} &54.53\scriptsize{$\pm$1.99} &88.70 $\pm$ 1.77 \\
    ACM &38.16\scriptsize{$\pm$0.86} &28.33\scriptsize{$\pm$0.82} &62.43\scriptsize{$\pm$1.45} & 24.56\scriptsize{$\pm$1.08} &18.56 \scriptsize{$\pm$0.76} & 57.41\scriptsize{$\pm$0.47} &24.22\scriptsize{$\pm$1.28} &18.96\scriptsize{$\pm$0.92} &51.26\scriptsize{$\pm$1.28} &68.66\scriptsize{$\pm$1.73} &56.28\scriptsize{$\pm$0.63} &90.65 $\pm$ 0.84 \\
    Oversampling &37.28\scriptsize{$\pm$2.19} &28.05\scriptsize{$\pm$1.71} &61.32\scriptsize{$\pm$2.85} & 23.78\scriptsize{$\pm$1.17} & 16.68\scriptsize{$\pm$ 1.05}&56.02 \scriptsize{$\pm$ 1.37}&22.11\scriptsize{$\pm$1.81} &17.15\scriptsize{$\pm$1.45}&50.39\scriptsize{$\pm$1.21} &66.00\scriptsize{$\pm$2.02} &55.52\scriptsize{$\pm$1.79} &89.14 $\pm$ 1.53 \\
    Re-weight &36.40\scriptsize{$\pm$1.36} &27.59\scriptsize{$\pm$1.25} &59.47\scriptsize{$\pm$1.61} &27.98\scriptsize{$\pm$1.34}&20.95\scriptsize{$\pm$ 1.19}&58.52\scriptsize{$\pm$1.38}&21.34\scriptsize{$\pm$1.85} &16.08\scriptsize{$\pm$1.70} &51.52\scriptsize{$\pm$1.58} &65.69\scriptsize{$\pm$1.35} &55.65\scriptsize{$\pm$1.06} &89.39 $\pm$1.53 \\
    DR-GCN &37.36\scriptsize{$\pm$2.85} &28.78\scriptsize{$\pm$2.44} &60.34\scriptsize{$\pm$1.63} &19.03\scriptsize{$\pm$0.75} &15.23\scriptsize{$\pm$0.46} &47.43\scriptsize{$\pm$0.49} &15.57\scriptsize{$\pm$1.34} &11.62\scriptsize{$\pm$1.34} &47.28\scriptsize{$\pm$0.82} &65.92\scriptsize{$\pm$1.64} &60.90\scriptsize{$\pm$1.25} &84.35 \scriptsize{$\pm$4.49}  \\
    ImGAGN  & 44.05\scriptsize{$\pm$0.75} & 33.21\scriptsize{$\pm$0.60} & 69.62\scriptsize{$\pm$0.16} & 21.23\scriptsize{$\pm$0.45} & 13.86\scriptsize{$\pm$0.46} & 51.81\scriptsize{$\pm$0.36} &18.86 \scriptsize{$\pm$0.72} & 13.82\scriptsize{$\pm$0.67} & 54.16\scriptsize{$\pm$0.24} &79.97\scriptsize{$\pm$1.42} &63.83\scriptsize{$\pm$1.06} &95.59 $\pm$0.44 \\
    GraphSMOTE  &36.92\scriptsize{$\pm$0.59} &27.43\scriptsize{$\pm$0.54} &61.13\scriptsize{$\pm$0.29} & 23.75\scriptsize{$\pm$0.41} &17.26 \scriptsize{$\pm$0.38} &53.37 \scriptsize{$\pm$0.21} &21.54\scriptsize{$\pm$1.73} &16.11\scriptsize{$\pm$1.71} &50.39\scriptsize{$\pm$0.28} &82.81\scriptsize{$\pm$0.59} &72.44\scriptsize{$\pm$1.29} & 96.49 $\pm$0.18\\
    GraphENS  & 31.43\scriptsize{$\pm$0.56} & 26.06\scriptsize{$\pm$0.52} & 64.37\scriptsize{$\pm$0.15} & 26.72\scriptsize{$\pm$0.27} & 18.96\scriptsize{$\pm$0.92} & 51.87\scriptsize{$\pm$0.08} & 26.80\scriptsize{$\pm$0.43} & 24.63\scriptsize{$\pm$0.55} & 55.95\scriptsize{$\pm$0.10} &89.68\scriptsize{$\pm$0.25} &87.22\scriptsize{$\pm$0.28} &98.90 \scriptsize{$\pm$0.04} \\
    \midrule
    \textbf{\name} &\textbf{49.01}\scriptsize{$\pm$1.24} &\textbf{48.29}\scriptsize{$\pm$0.25} &\textbf{77.07}\scriptsize{$\pm$0.87} & \textbf{30.20}\scriptsize{$\pm$1.02} &\textbf{26.53}\scriptsize{$\pm$0.14} &\textbf{61.41}\scriptsize{$\pm$0.97} & \textbf{27.89}\scriptsize{$\pm$0.56} &\textbf{26.07}\scriptsize{$\pm$0.28} &\textbf{57.64}\scriptsize{$\pm$1.04} &\textbf{91.56}\scriptsize{$\pm$0.72} &\textbf{90.43}\scriptsize{$\pm$0.41} & \textbf{99.43}\scriptsize{$\pm$0.25} \\ 
    \bottomrule
\end{tabular}
\caption{Comparision of \name with other baselines in semi-supervised setting (\textit{im\_ratio}=0.1).We report the averaged accuracy, F1-score and AUC-ROC with the standard
errors for 5 repetitions on six imitative imbalanced datasets for node classification.}
  \label{tab:semi}
\end{table*}
\vspace{-0.2cm}

\vspace{-2mm}
\section{Experiment}
In this section, we conduct extensive experiments on eight public datasets to evaluate the effectiveness of GraphSANN, which aim to answer five research questions: \textbf{RQ1}: How does \name perform compared to other baselines in imbalanced node classification on both homophilic and heterophilic graphs? \textbf{RQ2}: How effective is \name under different imbalance ratios? \textbf{RQ3}: How does each core component of \name contribute to the performance gain? \textbf{RQ4}: How do different hyper-parameter values affect the performance of \name? \textbf{RQ5}: Can \name learn effective node representation to separate different classes of nodes in the embedding space?
\vspace{-0.1cm}
\subsection{Experimental Setup}
\xhdr{Datasets} To thoroughly evaluate the performance of \name, we conduct experiments on eight benchmark datasets including six artificial imbalanced datasets and two genuine ones. Among the artificial datasets, Cora, Citeseer and Pubmed are three citation networks with high homophily, while Chameleon, Squirrel and Film are three Wikipedia networks with high heterophily. 3, 3, 2, 2, 2, and 2 classes are randomly selected as minority classes for these six datasets by down-sampling. Following \cite{zhao2021graphsmote}, all majority classes have 20 nodes while minority classes only have $20\ \times$ \textit{im\_ratio} nodes in the training set. For two Amazon product networks whose class distributions are genuinely imbalanced, we use their original class ratios. The detailed statistical information of the six datasets is summarized in Table \ref{tab:datasets}.

\xhdr{Baselines} We compare \name with eight state-of-the-art baselines for imbalanced node classification problem, including two vanilla models: \textit{GCN}~\cite{kipf2016semi} and \textit{ACM}~\cite{luan2022revisiting}; two generic class-imbalance methods: \textit{Oversampling} and \textit{Reweight}; and four network-specific methods: \textit{DR-GCN}~\cite{shi2020multi}, \textit{ImGAGN}~\cite{qu2021imgagn}, \textit{GraphSMOTE}~\cite{zhao2021graphsmote} and \textit{GraphENS}~\cite{park2021graphens}. Please refer to Appendix \ref{sub:setup} for detailed descriptions of each baseline.

\xhdr{Evaluation Metrics} Following existing works~\cite{zhao2021graphsmote} in evaluating imbalanced classification, three evaluation metrics are adopted in this paper: Accuracy, AUC-ROC, and Macro-F1,
where both AUC-ROC and Macro-F1 are reported by averaging the metrics over each class. 

\xhdr{Parameter settings}
The following hyper-parameters are set for our model in all the datasets. Layer number $L=2$ with hidden dimensions 64 and 32 for both edge generation and node classification. Adam optimizer with learning rate $lr=0.001$ for homophilic graphs and 0.01 for heterophilic graphs. Dropout rate $\gamma=0.7$. $\textit{Epochs}=2000$ with early stop strategy. \textit{Weight\_decay}$=5\mathrm{e}{-4}$. Hyper-parameters $\kappa=1.05$, $\omega=0.3$. Initial hop $h=2$, threshold $\eta=0.5$ and loss weight $\lambda=1\mathrm{e}{-6}$. Sampling ratio of candidate edges $\xi=0.3$. Over-sampling scale $\zeta=1.0$.


\vspace{-0.2cm}
\subsection{Imbalanced Node Classification (RQ1)}
To answer RQ1, we compare the node classification performance of \name with other baselines across all eight datasets and report the average performances along with standard deviations of each metric. Table \ref{tab:semi} shows the node classification results for six imitative datasets and two genuine datasets. From the table, we can observe that: (1) \name outperforms all the other baselines by all metrics on all eight datasets. This indicates our proposed model consistently acquires better performance on either homophilic or heterophilic networks.
(2) On three heterophilic datasets, most class-imbalance baselines only acquire slightly better, or even worse performance (e.g. \textit{DR-GCN} on Film and \textit{GraphSMOTE} on Squirrel) than vanilla models i.e., \textit{GCN} or \textit{ACM}. This is because these baselines rely on homophilic assumption and generate synthetic edges based on feature similarity. Thus they perform poorly on heterophilic graphs whose edges link nodes with dissimilar features.
GraphSANN, however, thanks to the adaptive subgraph extractor and multi-filter encoder blocks, can discriminatively aggregate similar node features to generate heterophilic edges, and thus achieve significant performance gains over the baselines.
(3) On genuine datasets, compared to the most competitive baseline \textit{GraphENS}, \name still acquires $4.5\%$ and $3.6\%$ performance gains w.r.t. F1-score on Amazon-Computers and Amazon-Photo, respectively.

\begin{figure*}[t]
  \centering
  \includegraphics[scale=0.5]{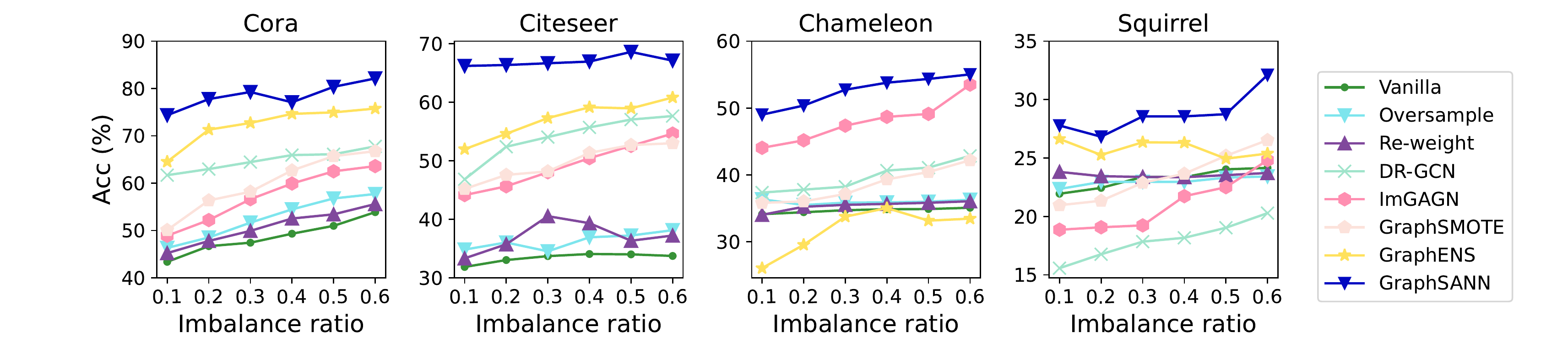}
  \vspace{-2mm}
  \caption{Node classification results under different imbalance ratios.}\label{fig:im_ratio}
\end{figure*}

\vspace{-0.2cm}
\subsection{Influence of Imbalance Ratio (RQ2)}
To answer RQ2, in this subsection, we further compare the performance of \name with other baselines under different $im\_ratios$. 
The imbalance ratio varies from 0.1 to 0.6. Each experiment is repeated 5 times and the average results are reported in Figure \ref{fig:im_ratio}. From Figure \ref{fig:im_ratio}, we can observe that (1) \name consistently outperforms other baselines across all the imbalance ratios on all the datasets, this demonstrates the generalization and robustness of our model under different imbalanced scenarios. (2) Generally, \name has more significant performance improvement over other baselines under more extreme imbalance ratios. As imbalance ratio increases, the datasets become more balanced, which offsets the effects brought by node/edge augmentation. 

 \vspace{-0.1cm}
\subsection{Ablation Study (RQ3)}
 \vspace{-0.1cm}
To further investigate the contribution of each component of GraphSANN, we perform an ablation study and report the results in Table \ref{tab:ablation}. $w/o\ \textit{UFM}$ replaces the unified feature mixer component with simple SMOTE strategy~\cite{chawla2002smote}; $w/o\ \textit{ASE}$ replaces the adaptive subgraph extractor with fixed 2-hop neighbors of target edge to form subgraphs; and $w/o\ \textit{MSE}$ replaces multi-filter subgraph encoder component with a raw GCN. \name represents the full model with all the components available. From this table, we can observe that: (1) All three components contribute to the performance improvement of \name; (2) Adaptive subgraph extractor and multi-filter encoder exhibit crucial effects on heterophilic networks in view of sharp performance drops between $w/o\ \textit{ASE}$ and \name and between $w/o\ \textit{MSE}$ and \name on Chameleon.

\begin{table}[t]\small
  \setlength\tabcolsep{2.5pt} 
  \centering
  \begin{tabular}{l|ccc|ccc}
    \toprule
    \multirow{2}{*}{\textbf{Method}} &\multicolumn{3}{c|}{\textbf{Cora}} &\multicolumn{3}{c}{\textbf{Chameleon}} \\
     & ACC & F1 & AUC & ACC & F1 & AUC \\
    \midrule
    $w/o\ \textit{UFM}$ & 68.32 &66.45 &93.87 &43.25 &42.68 & 69.61  \\
    $w/o\ \textit{ASE}$ & 75.14 & 73.21 &94.21 &45.65 &43.95 & 72.25  \\
    $w/o\ \textit{MSE}$ & 76.20 & 73.52 &94.33 &42.33&40.75 & 70.39 \\
    \textbf{\name} & \textbf{77.73} &\textbf{74.94} &\textbf{95.59}  &\textbf{49.01} &\textbf{48.29} &\textbf{77.07}  \\
    \bottomrule
\end{tabular}
\caption{Ablation study results.}
\label{tab:ablation}
\end{table}

\vspace{-0.2cm}
\subsection{Parameter Sensitivity Study (RQ4)}
\begin{figure}[t]
  \centering
  \vspace{-3mm}
  \subfigure[Dropout rate]{
  \includegraphics[scale=0.24]{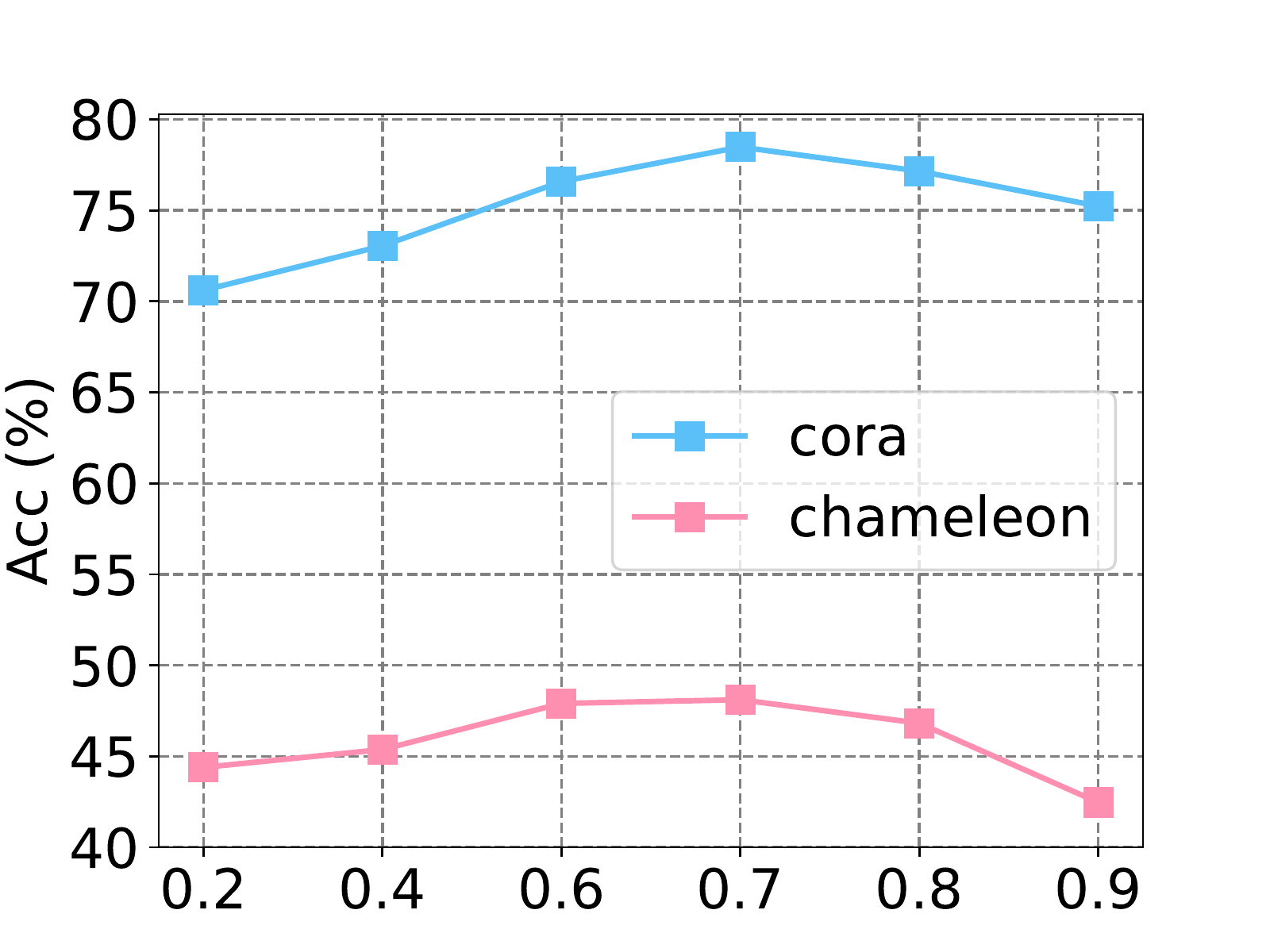}
  }
  \subfigure[Sampling ratio]{
  \includegraphics[scale=0.24]{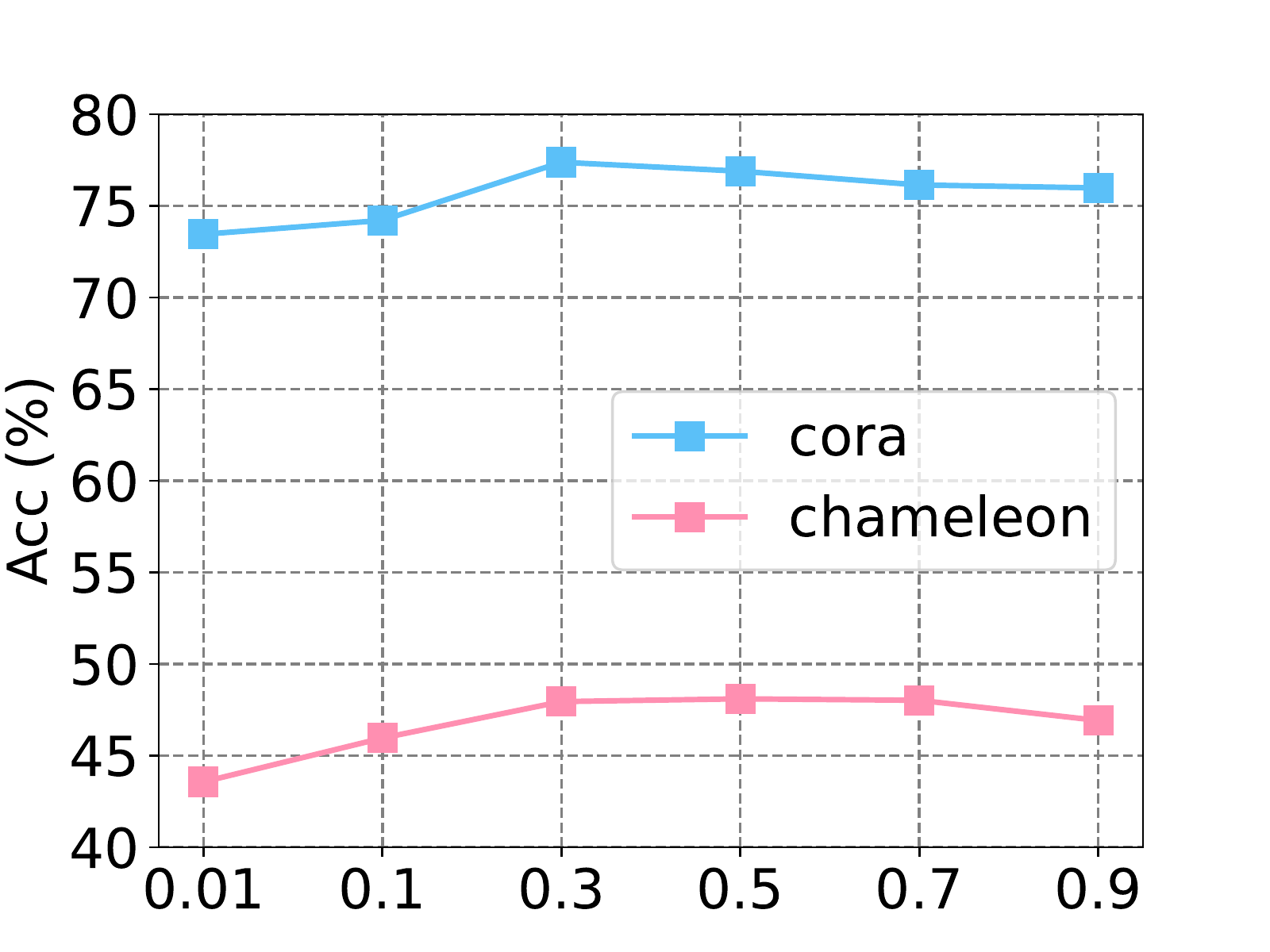}
  }
  \vspace{-2mm}
  \caption{Hyper-parameter sensitivity analysis of dropout rate $\gamma$ and candidate edge sampling ratio $\xi$.} \label{fig:hyperpara}
\end{figure}

\vspace{-0.1cm}
In this subsection, we investigate the impact of two crucial hyper-parameters, i.e., dropout rate $\gamma$ of adaptive classifier and sampling ratio $\xi$ of candidate synthetic edges on model performance. We vary $\gamma$ from 0.1 to 0.9 with step size 0.1 and vary $\xi$ from 0.01 to 0.9. The experiments are conducted on both Cora and Chameleon and test accuracy curves are shown in Figure \ref{fig:hyperpara}. 
From Figure \ref{fig:hyperpara}, we can observe that: (1) As $\gamma$ increases, model performance gradually rises and reaches peak values when $\gamma$ reaches 0.7 on both datasets. Then, as dropout rate keeps increasing, the performance gradually drops. 
(2) The performance decreases as the sampling ratio is under 0.3 or over 0.5. Our explanation is that the original edge distribution is not sufficiently simulated at low $\xi$, while a high $\xi$ impedes feature aggregations by introducing too many noisy edges, resulting in performance degradation.

\begin{figure}[t]
  \centering
  \vspace{-3mm}
  \subfigure[GraphENS]{
  \includegraphics[scale=0.19]{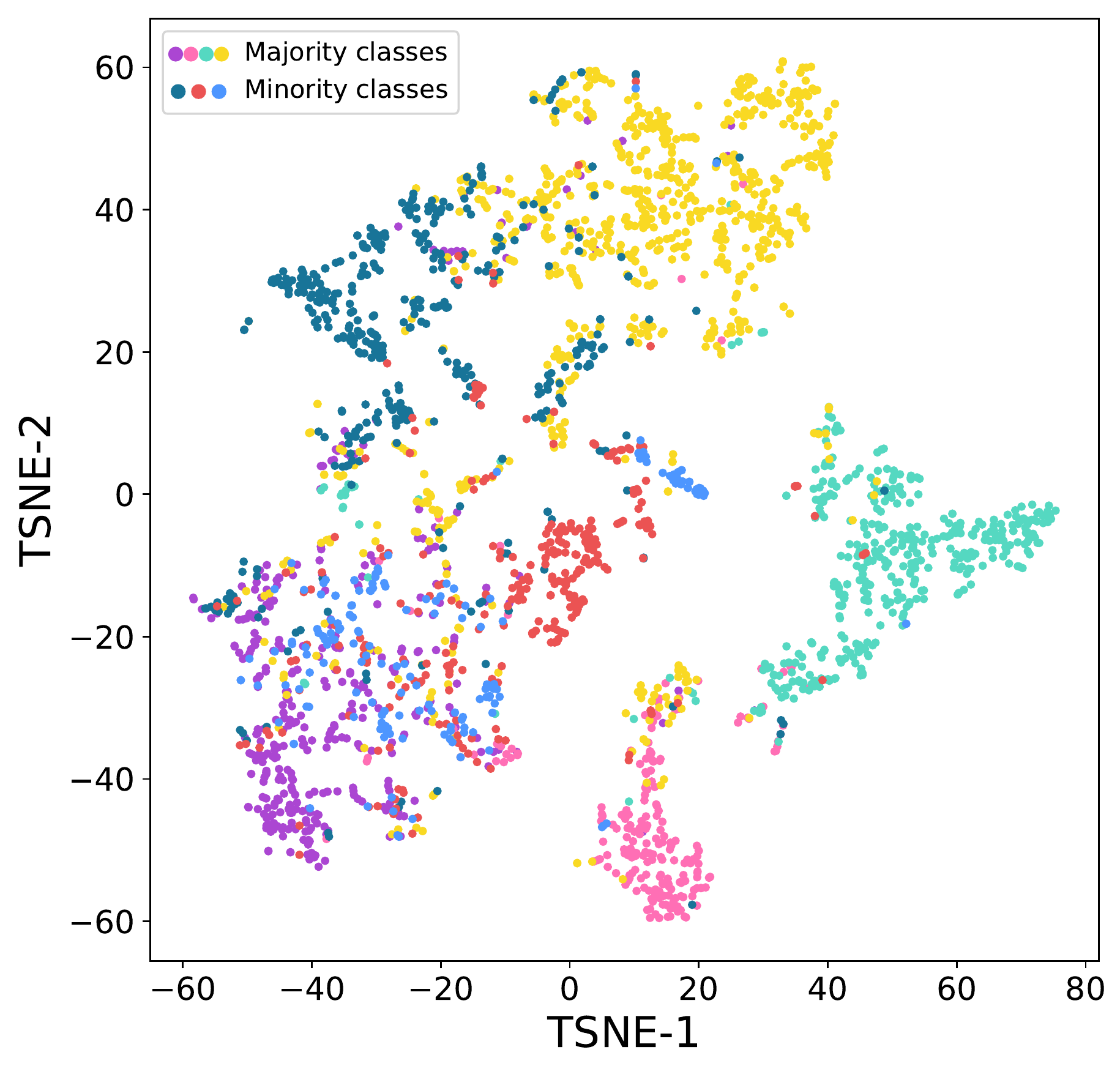}
  }\hspace{-2mm}
  \subfigure[\name]{
  \includegraphics[scale=0.19]{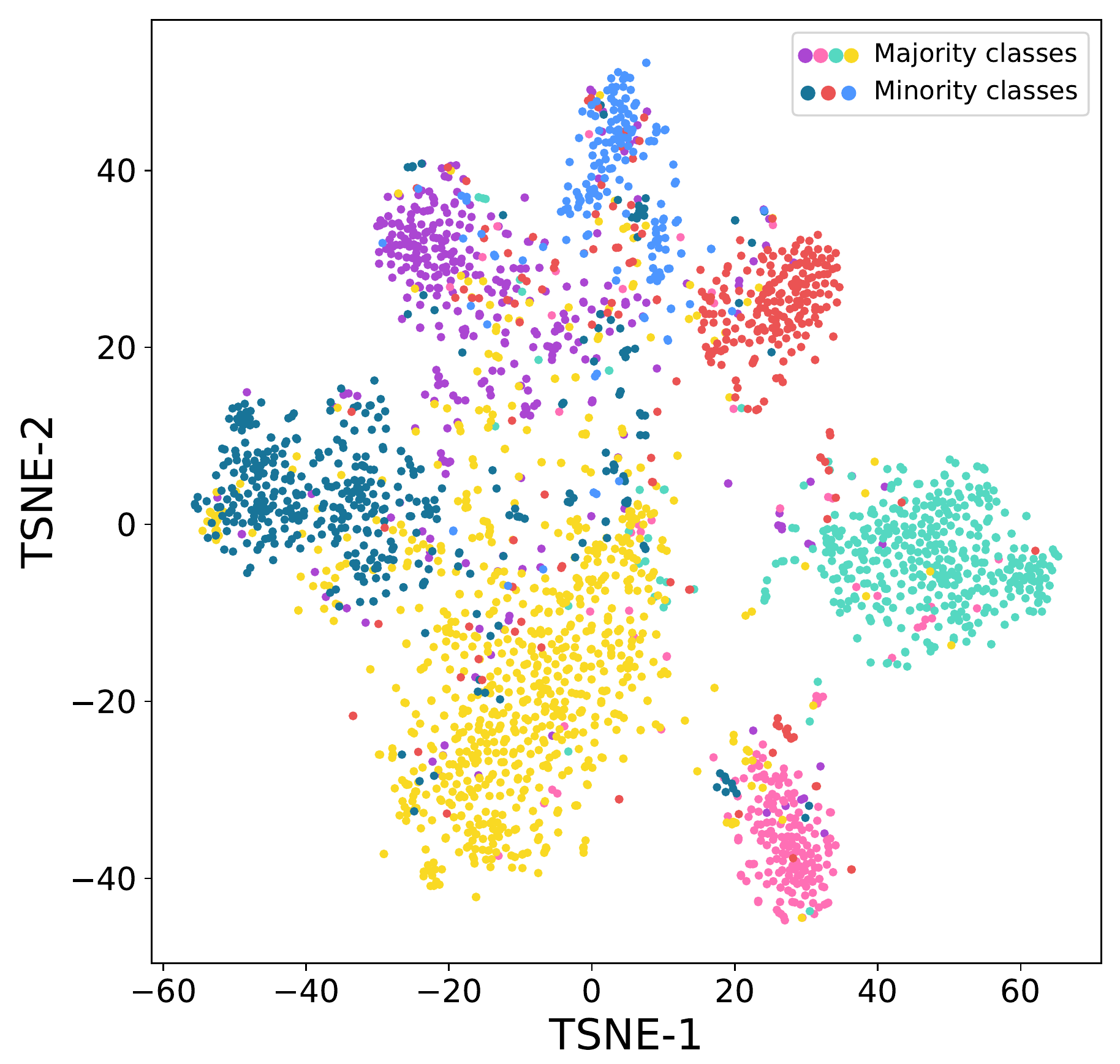}
  }
  \vspace{-2mm}
  \caption{Visualization of \name and GraphENS.} \label{fig:visual}
\end{figure}

\vspace{-0.2cm}
\subsection{Visualization (RQ5)}
In this subsection, we project the latent node embeddings of \name and the most competitive baseline \textit{GraphENS} on Cora into two-dimensional space using t-SNE~\cite{van2008visualizing} and color the nodes based on their class labels. As shown in Figure \ref{fig:visual}, we can observe that the minority class representations of \textit{GraphENS} (e.g. blue and dark cyan dots) are hard to be distinguished and have large mixed areas with majority clusters, while node representations of \name are clustered tightly together with clear boundaries for both majority and minority classes. This proves the superiority of \name in terms of embedding quality for separating different classes despite the class-imbalance problem.

\vspace{-0.2cm}
\section{Conclusion}
\vspace{-0.1cm}
In this paper, we design a novel \name for imbalanced node classification on both homophilic and heterophilic graphs. The elaborately designed three components within it can unifiedly interpolate synthetic nodes, adaptively extracts surrounding subgraphs of candidate synthetic edges, and discriminatively encode them to predict the existence of synthetic edges. Extensive experiments on eight benchmark datasets have demonstrated the superiority of GraphSANN.

\section*{Acknowledgments}
This work is partially supported by The National Key Research and Development Program of China (Grant No. 2020AAA0108504), Australian Research Council Future Fellowship (Grant No. FT210100624), Discovery Project (Grant No. DP190101985), and Discovery Early Career Research Award (Grant No.DE200101465).

\bibliographystyle{named}
\bibliography{Imbalance}

\clearpage
\newpage

\section{Appendix}\label{sec:appen}
\subsection{Notations}
\begin{table}[h]
  \renewcommand\arraystretch{1.1}
  \centering
  \begin{tabular}{l|l}
    \toprule
    \textbf{Notations} & \textbf{Definitions} \\
    \midrule
    $\mathcal{G}$ & Attributed graph. \\
    $\mathcal{V}$ & Node set for nodes in $\mathcal{G}$. \\
    $\mathcal{V}_{minor}$ & Node set for nodes with minority classes. \\
    $\mathcal{V}_c$ & Node set for nodes with class $c$. \\
    $\mathcal{V}_{syn}$ & Node set for synthetic nodes. \\
    $\mathcal{E}$ & Edge set for edges in $\mathcal{G}$. \\
    $\mathcal{E}_{syn}$ & Edge set for synthetic edges. \\
    $C$ & Number of node classes. \\
    $\mathcal{C}$ & Node class set. \\
    $\mathcal{C}_M$ & Minority node class set. \\
    $\hat{\psi}_{st}$ & Similarity between node features. \\
    $S_{pair}$ & Sampled node pair set. \\
    $\mathcal{N}_h(\cdot)$ & h-hop neighbors of a node. \\
    $\mathbf{X}$ & Node feature matrix. \\
    $\mathbf{Y}$ & One-hot label matrix for nodes. \\
    $\mathbf{h}^{(l)}$ & The l-th layer node feature. \\
    $\mathbf{D}_t$ & Integrated Gradient matrix.\\
    $\mathbf{W}_p$, $\mathbf{W}_k$ & Weight matrices. \\
    $\mathbf{W}^{(l)}_L$ & Weight matrices of low-pass messages. \\
    $\mathbf{W}^{(l)}_{H}$ & Weight matrices of high-pass messages. \\
    $\mathbf{W}^{(l)}_{I}$ & Weight matrices of identity messages. \\
    $\mathbf{W}_{pool}$ & Pooling weight matrix. \\
    $\mathbf{M}_t$ & Mask vector.\\
    $\boldsymbol{\alpha}^{(l)}_{i,j}$ & Multi-pass weight vector in the $l$-th layer. \\
    \bottomrule
  \end{tabular}
  \caption{Summary of Notations.}\label{table:nota}
\end{table}

\subsection{Saturation and Thresholding Problems}
\begin{figure}[h]
    \subfigure[Saturation problem]{
    \includegraphics[scale=0.9]{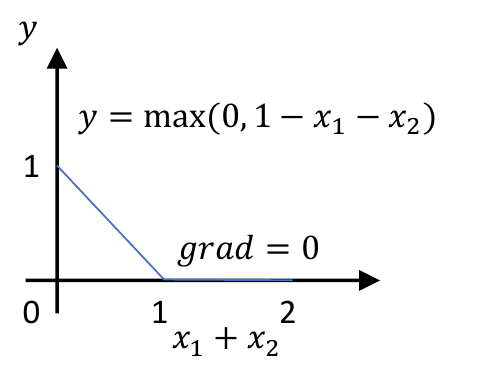}
    }\subfigure[Thresholding problem]{
    \includegraphics[scale=0.9]{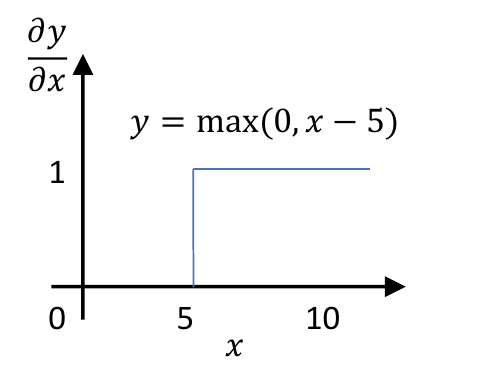}
    }
    \caption{Illustrations of saturation and thresholding problems for gradient-based importance score method.}
    \label{fig:gradient}
\end{figure}

\cite{park2021graphens} utilizes the gradient of node classification loss w.r.t. input features to calculate the importance of each node attribute. Despite its simplicity, this approach suffers from both saturation and thresholding problems, which we will state in detail in this subsection. First of all, the gradient-based approach cannot model saturation situations. Consider a simple network $y=max(0,1-x_1-x_2)$ with $x_1$  and $x_2$ as inputs and a rectified linear unit (ReLU) as activation function. As figure \ref{fig:gradient}(a) shows, at the point of $x_1=1$ and $x_2=1$, perturbing either $x_1$ or $x_2$ from 1 to 0 will not change the output, and thus the gradient of output w.r.t. the inputs will stay 0 as long as $x_1+x_2>1$. This example shows that gradients could underestimate the importance of features that saturate their contributions to the output~\cite{shrikumar2017learning}. Second, the gradient-based approach suffers from thresholding problem: Consider a rectified linear unit with a bias of -5: $y=max(0, x-5)$,  as figure \ref{fig:gradient}(b) shows, the gradient has a discontinuity at $x=5$, which causes sudden jumps in the importance score over infinitesimal changes in the inputs. Therefore, the gradient-based method has drawbacks to compute continuous and stable importance scores.

In view of that, instead of computing the gradients only at the current value of the input, we integrate the gradients as the inputs scale up from all zeros to their current values~\cite{sundararajan2017axiomatic}. This addresses the saturation and thresholding problems and produces more stable importance scores of input features.
\subsection{Integrated Gradient Computation}\label{sub:int}
As illustrated by \cite{sundararajan2017axiomatic}, the integral of integrated gradient shown in Eq (\ref{eq:integral}) can be approximately calculated by summation operation, which can be formed as follows:
\begin{equation}
    \text{IG}_i(\mathbf{x})\approx \frac{1}{S}\sum_{s=1}^S\frac{\partial\mathcal{L}_{cls}(\frac{s}{S}\mathbf{x}, \mathbf{y})}{\partial\mathbf{x}_i}\mathbf{x}_i,
\end{equation}
where $S$ is the number of steps in the Riemann approximation of the integral. The error between the Riemann sum and the integral satisfies $error \leq M_1\frac{\mathbf{x}_i}{S}$, where $M_1$ is the upper bound for $\frac{\partial \mathcal{L}_{cls}(t\mathbf{x},\mathbf{y})}{\partial\mathbf{x}_i}$ over $t\in[0,1]$. To balance the approximation precision and time complexity, we choose the step number $S$ as 50, which can guarantee the approximation error is within $5\%$ in an acceptable computational cost.

\subsection{Complexity Analysis}
According to Algorithm \ref{alg:1}, the computational cost of \name mainly comes from two parts: integrated gradient-based feature mixup and multi-filter subgraph encoder. Specifically, as introduced by \ref{sub:int}, given the steps of Riemann approximation $S$ and oversampling ratio $\zeta$, the time complexity of integrated gradient-based feature mixup is $\mathcal{O}\bigl( S|\mathcal{C}_M||\mathcal{V}_m|\zeta\bigr)$, where $\mathcal{C}_M$ and $\mathcal{V}_m$ is the minority class set and minority node-set, respectively. Since the sizes of minority node sets are usually very small, the complexity of this part is not very high. Let $|\mathcal{V}_e|$ and $|\mathcal{E}_e|$ be the average node and edge numbers of a subgraph $\mathcal{G}_e$, $F$ and $F'$ be the input and output dimension of a multi-filter encoder layer, and $K$ be the number of extracted subgraphs. The time complexity of a one-layer multi-filter subgraph encoder can be denoted as $\mathcal{O}\bigl(3K(|\mathcal{V}_e|FF'+|\mathcal{E}_e|F')\bigr)$. Compared with feature similarity-based edge generation models~\cite{zhao2021graphsmote,park2021graphens}, \name introduces extra computational cost for extracting and encoding a subgraph for each candidate edge. However, for large graphs which cannot directly fit into GPU memory, a mini-batch training strategy has to be used for both feature similarity-based models and ours, which results in similar computational costs.

\subsection{Experimental Setup}\label{sub:setup}
\xhdr{Dataset} The detailed settings of all eight datasets are described as follows:
\begin{itemize}[leftmargin=*]
    \item \textbf{Artificial imbalanced datasets}:
    Since the class distribution of the citation networks and Wikipedia networks are relatively balanced, we use an imitative imbalanced setting. Specifically, 3, 3, 2, 2, 2, and 2 classes are randomly selected as minority classes and down-sampled for each dataset, respectively. In the semi-supervised setting, following \cite{zhao2021graphsmote}, all majority classes have 20 nodes in the training set while minority classes only have $20\ \times$ \textit{im\_ratio} nodes. We vary \textit{im\_ratio} to analyze the performance of \name under various imbalanced scenarios. In the supervised learning setting, we follow \cite{qu2021imgagn} to randomly split the training, validation, and testing set in a ratio of 7:1:2. In the training set, the \textit{im\_ratio} between minority classes and majority classes are set as 0.1. Note that in both semi-supervised and supervised settings, we sample the same number of nodes from each class for validation/test sets 
    \item \textbf{Genuine imbalanced datasets}:
    For the Amazon product networks whose class distributions are genuinely imbalanced, we use their original class ratios. For semi-supervised learning, we set the total labeled training nodes as 50 and 30 for Amazon-Computers and Amazon-Photo, respectively. 10$\%$ and 20$\%$ of the total nodes are selected as validation and testing sets, respectively. For the supervised setting, we also randomly split the training, validation, and testing set in a ratio of 7:1:2. 
\end{itemize}

\xhdr{Baselines} We choose seven state-of-the-art imbalanced node classification methods as baselines, which are described in detail as follows:
\begin{itemize}[leftmargin=*]
    \item \textbf{GCN}~\cite{kipf2016semi}: Original implementation of a homophilic GNN without additional tricks dealing with class imbalance problem.
    \item \textbf{ACM}~\cite{luan2022revisiting}: Original implementation of a heterophilic GNN without additional tricks dealing with class imbalance problem.
    \item \textbf{Oversampling}: A classical imbalanced learning approach by repeating samples from minority classes. Following \cite{chawla2002smote}, we implement it by duplicating minority nodes in node embedding space.
    \item \textbf{Reweight}~\cite{yuan2012sampling+}: A cost-sensitive approach that assigns higher loss weights to samples from minority classes. Here we select a 2-layer GCN~\cite{kipf2016semi} as the backbone model for the first three baselines.
    \item \textbf{DR-GCN}~\cite{shi2020multi}: A GCN-based imbalanced network embedding method that uses class-conditional adversarial training to enhance the separation of different classes. 
    \item \textbf{ImGAGN}~\cite{qu2021imgagn}: A generative adversarial imbalanced network embedding model which utilizes an MLP as a graph generator and a GCN as a node discriminator. It is originally designed for binary node classification and we extend it to the multi-class case.
    \item \textbf{GraphSMOTE}~\cite{zhao2021graphsmote}: By extending SMOTE\cite{chawla2002smote} to graph scenario, Graph-SMOTE can generate synthetic nodes and link them to existing nodes through a pre-trained edge generator.
    \item \textbf{GraphENS}~\cite{park2021graphens}: An augmentation-based method that synthesizes the whole ego network for minority classes by combining diverse ego networks based on similarity. It is reported to have acquired SOTA performances on multiple imbalanced node classification datasets.
\end{itemize}
\xhdr{Implementation Details} The proposed \name is implemented in PyTorch and optimized by Adam Optimizer~\cite{kingma2014adam}. The model is trained and tested in a 24 GB Titan RTX GPU. Specifically, layer number $L$ of the multi-filter encoder is set as 2 with hidden dimensions 64 and 32 for both edge generation and node classification. We grid search for the learning rate in $\{0.0001, 0.0005, 0.001,0.005,0.01,0.05,0.1\}$, dropout rate $\gamma$ in $\{0.1,0.2,0.3,0.4,0.5,0.6,0.7,0.8,0.9\}$, weight decay in $\{1\mathrm{e}{-5},5\mathrm{e}{-4},1\mathrm{e}{-4},5\mathrm{e}{-3},1\mathrm{e}{-3},5\mathrm{e}{-2},1\mathrm{e}{-2},1\mathrm{e}{-1}\}$ and sampling ratio $\xi$ in $\{0.01,0.1,0.3,0.5,0.7,0.9\}$. Hyper-parameters above are selected according to the optimal performances of models on validation sets. Other hyper-parameters are selected based on previous works: $\kappa=1.05$, $\omega=0.3$, initial hop $h=2$, threshold $\eta=0.5$, loss weight $\lambda=1\mathrm{e}{-6}$, and over-sampling scale $\zeta=1.0$. We set $\textit{epochs}=2000$ and stop early if the performance doesn’t increase for 5 consecutive epochs on the validation set. Mini-batch strategy is applied for the training stages of both edge generation and node classification and the batch size is set as 32 for all datasets.

\begin{table}\small
  \setlength\tabcolsep{1.7pt} 
  \centering
  \begin{tabular}{l|cc|cc}
    \toprule
    \multirow{2}{*}{\textbf{Method}} & \multicolumn{2}{c}{\textbf{Amazon-Computers}} & \multicolumn{2}{c}{\textbf{Amazon-Photo}} \\
     & ACC & F1 & ACC & F1 \\
    \midrule
    GCN &64.61\scriptsize{$\pm$0.05} &45.96\scriptsize{$\pm$0.11} &82.38\scriptsize{$\pm$1.54} &68.01\scriptsize{$\pm$3.51} \\
    ACM &67.48\scriptsize{$\pm$0.43} &50.71\scriptsize{$\pm$0.56} &83.46\scriptsize{$\pm$0.89} &70.45\scriptsize{$\pm$1.68} \\
    Oversampling &66.36\scriptsize{$\pm$0.05} &49.50\scriptsize{$\pm$0.10} &81.22\scriptsize{$\pm$2.64} &68.46\scriptsize{$\pm$1.46} \\
    Re-weight &62.06\scriptsize{$\pm$0.09} &42.18\scriptsize{$\pm$0.17} &82.61\scriptsize{$\pm$1.21} &69.27\scriptsize{$\pm$1.39} \\
    DR-GCN &52.25\scriptsize{$\pm$1.03} &39.52\scriptsize{$\pm$1.83} &71.93\scriptsize{$\pm$1.75} &64.09\scriptsize{$\pm$1.63} \\
    ImGAGN &61.97\scriptsize{$\pm$0.20} &48.44\scriptsize{$\pm$0.10} &75.48\scriptsize{$\pm$0.88} &60.13\scriptsize{$\pm$1.02} \\
    GraphSMOTE & 75.48\scriptsize{$\pm$0.26}&69.68\scriptsize{$\pm$0.39} &87.52\scriptsize{$\pm$0.44} &80.19\scriptsize{$\pm$0.69} \\
    GraphENS &87.82\scriptsize{$\pm$0.24} &86.59\scriptsize{$\pm$0.26} &94.37\scriptsize{$\pm$0.16} &93.15\scriptsize{$\pm$0.24} \\
    \midrule
    \textbf{\name} &\textbf{89.42}\scriptsize{$\pm$0.84} &\textbf{87.71}\scriptsize{$\pm$0.45} &\textbf{96.41}\scriptsize{$\pm$0.66} &\textbf{95.78}\scriptsize{$\pm$0.82} \\
    \bottomrule
\end{tabular}
\caption{Node classification results on two genuine imbalanced datasets under supervised training setting in 5 repetitions.} 
\label{tab:super-gen}
\end{table}

\begin{table*}[!h]\small
  \setlength\tabcolsep{2.2pt} 
  \centering
  \begin{tabular}{l|ccc|ccc|ccc}
    \toprule
    \textbf{Method}&\multicolumn{3}{c|}{\textbf{Cora-supervised}} &\multicolumn{3}{c|}{\textbf{Chameleon-supervised}} & \multicolumn{3}{c}{\textbf{Squirrel-supervised}}\\ 
    \textbf{$\mathcal{H}_{edge}$}&\multicolumn{3}{c|}{0.7362} &\multicolumn{3}{c|}{0.2795} & \multicolumn{3}{c}{0.2416}\\ 
    \midrule
    Metrics($\%$)& ACC & F1 & AUC & ACC & F1 & AUC & ACC & F1 & AUC\\
    \midrule
    GCN & 63.23\scriptsize{$\pm$0.37} & 49.09\scriptsize{$\pm$0.28} & 89.16\scriptsize{$\pm$0.53} & 56.58\scriptsize{$\pm$0.41} & 56.12\scriptsize{$\pm$1.52} & 82.14\scriptsize{$\pm$0.31} & 33.08\scriptsize{$\pm$ 0.42}& 31.59\scriptsize{$\pm$0.82} & 66.56\scriptsize{$\pm$0.63}  \\
     ACM & 64.58\scriptsize{$\pm$0.92} & 51.44\scriptsize{$\pm$0.43} & 91.56\scriptsize{$\pm$0.73} & 59.13\scriptsize{$\pm$0.33} & 59.25\scriptsize{$\pm$1.03} & 83.23\scriptsize{$\pm$0.45} & 36.48\scriptsize{$\pm$ 0.56}& 35.24\scriptsize{$\pm$0.55} & 67.89\scriptsize{$\pm$0.83}  \\
    Oversampling & 67.27\scriptsize{$\pm$0.49} & 55.51\scriptsize{$\pm$0.64} & 92.32\scriptsize{$\pm$0.32} & 59.65\scriptsize{$\pm$1.89} & 58.99\scriptsize{$\pm$1.87} & 82.74\scriptsize{$\pm$1.02} & 37.69\scriptsize{$\pm$0.34} &36.83\scriptsize{$\pm$0.18} & 67.38\scriptsize{$\pm$0.84} \\
    Re-weight  & 65.07\scriptsize{$\pm$0.23} & 54.21\scriptsize{$\pm$0.61} & 92.81\scriptsize{$\pm$0.56} & 58.77\scriptsize{$\pm$0.70} & 58.06\scriptsize{$\pm$0.74} & 82.27\scriptsize{$\pm$0.09} & 38.27\scriptsize{$\pm$0.99} & 37.77\scriptsize{$\pm$1.33} & 68.12\scriptsize{$\pm$0.31} \\
    DR-GCN & 73.53\scriptsize{$\pm$1.17} & 71.58\scriptsize{$\pm$1.24} & 91.62\scriptsize{$\pm$0.57} & 39.25\scriptsize{$\pm$0.99} & 34.04\scriptsize{$\pm$0.82} & 61.69\scriptsize{$\pm$0.62} & 20.59\scriptsize{$\pm$0.91} & 12.08\scriptsize{$\pm$1.49} & 51.12\scriptsize{$\pm$0.52} \\
    ImGAGN & 68.71\scriptsize{$\pm$1.42} & 64.06\scriptsize{$\pm$1.67} & 85.48\scriptsize{$\pm$1.33} & 44.61\scriptsize{$\pm$0.65} & 35.59\scriptsize{$\pm$0.42} & 79.69\scriptsize{$\pm$0.27} & 28.17\scriptsize{$\pm$0.62} & 21.14\scriptsize{$\pm$0.58} & 62.06\scriptsize{$\pm$0.74} \\
    GraphSMOTE & 69.49\scriptsize{$\pm$0.13} & 63.03\scriptsize{$\pm$0.99} & 92.26\scriptsize{$\pm$0.30} & 61.40\scriptsize{$\pm$1.71} & 59.48\scriptsize{$\pm$1.78} & 85.54\scriptsize{$\pm$0.42} & 44.04\scriptsize{$\pm$1.73} & 43.45\scriptsize{$\pm$1.72} & 72.52\scriptsize{$\pm$0.28} \\
    GraphENS & 78.08\scriptsize{$\pm$0.55} & 74.39\scriptsize{$\pm$0.77} & 93.22\scriptsize{$\pm$0.23} & 44.78\scriptsize{$\pm$0.76} & 44.63\scriptsize{$\pm$0.66} & 73.73\scriptsize{$\pm$0.15} & 31.92\scriptsize{$\pm$0.24} & 31.46\scriptsize{$\pm$0.29} & 63.04\scriptsize{$\pm$0.13} \\
    \midrule
    \textbf{\name}& \textbf{83.75}\scriptsize{$\pm$0.15} & \textbf{82.55}\scriptsize{$\pm$0.19} & \textbf{95.81}\scriptsize{$\pm$0.11} & \textbf{65.96}\scriptsize{$\pm$1.05} & \textbf{66.18}\scriptsize{$\pm$0.94} & \textbf{86.98}\scriptsize{$\pm$ 0.25}& \textbf{46.53}\scriptsize{$\pm$0.31} & \textbf{46.76}\scriptsize{$\pm$0.26} & \textbf{76.72}\scriptsize{$\pm$0.46} \\ 
    \bottomrule
\end{tabular}
\caption{Comparision of our method with other baselines under supervised training setting ($im\_ratio$=0.1) on three artificial imbalanced datasets. We report the averaged accuracy, F1-score, and AUC-ROC with the standard errors for 5 repetitions.} 
  \label{tab:super-art}
\end{table*}

\subsection{Additional Experimental Results}
\xhdr{Supervised Node Classification} We also conduct node classification under supervised setting that is introduced in subsection \ref{sub:setup}. We report the test results on two genuine imbalanced datasets: Amazon-Computers and Amazon-Photo, and three artificial imbalanced datasets: Cora, Chameleon and Squirrel in Table \ref{tab:super-gen} and Table \ref{tab:super-art}, respectively. We can observe that due to introducing more labeled nodes for both majority and minority classes in the training set, almost all the models acquire better performances under supervised setting than under semi-supervised setting. Despite that, \name also outperforms all the other baselines on all the datasets under supervised setting, which again demonstrate the effectiveness of our model.

\begin{figure}[tb]
  \centering
  \hspace{-4mm}\includegraphics[scale=0.33]{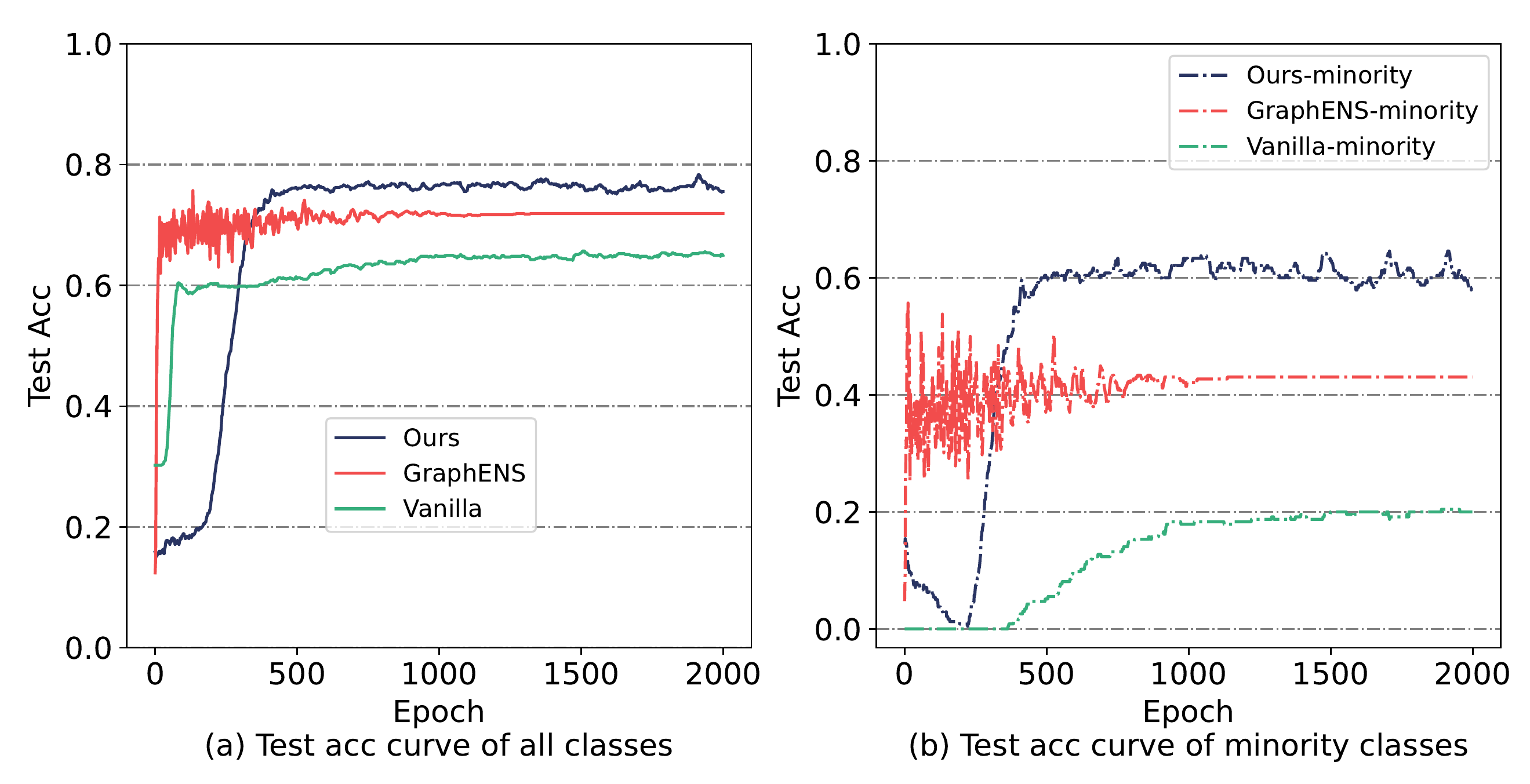}
  \caption{The learning curves of \name and two baselines.}\label{fig:test_curve}
\end{figure}

\begin{figure}[ht]
  \centering
  \subfigure[Vanilla GCN]{
  \includegraphics[scale=0.2]{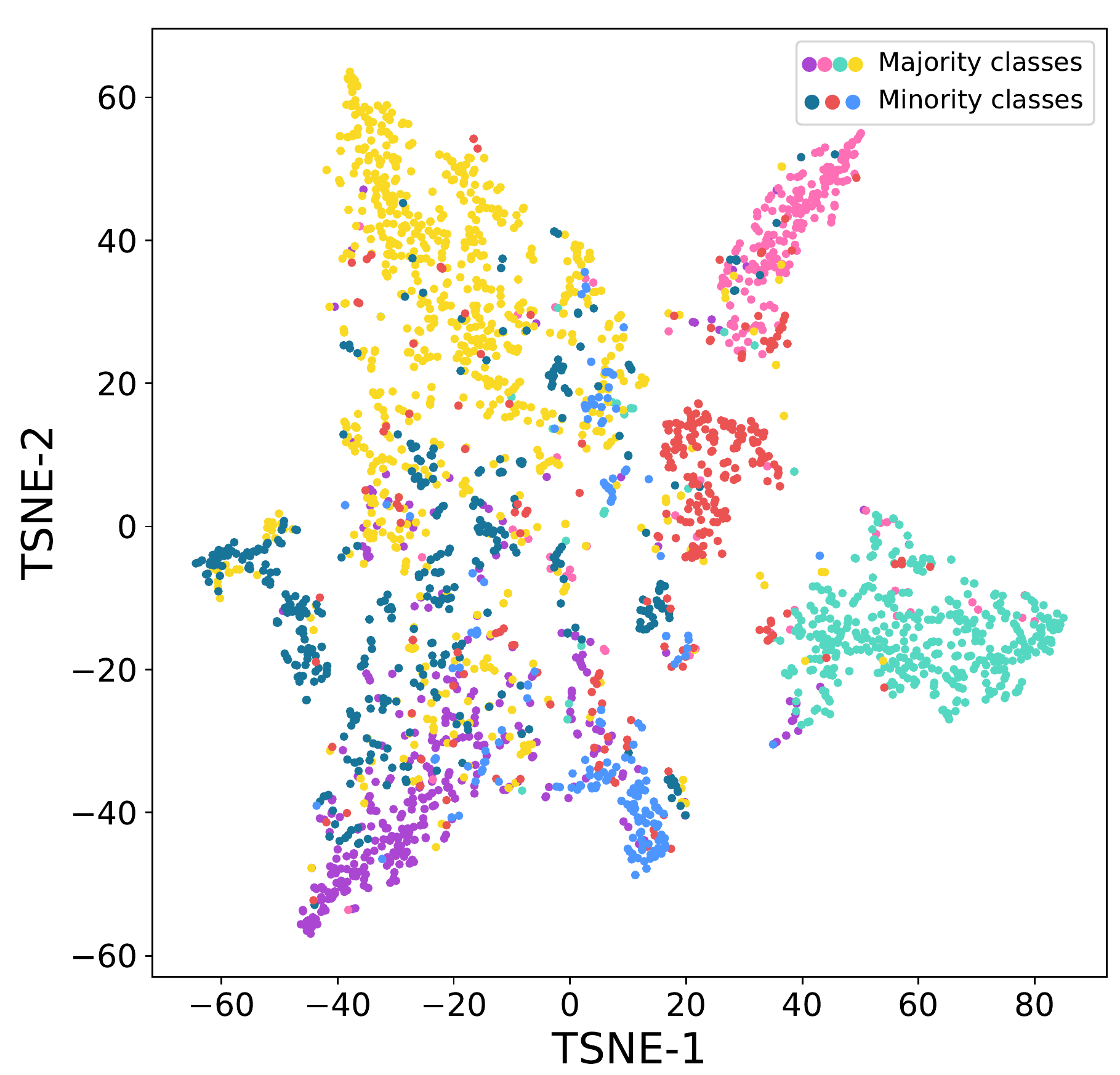}
  }\hspace{-2mm}
  \subfigure[ImGAGN]{
  \includegraphics[scale=0.2]{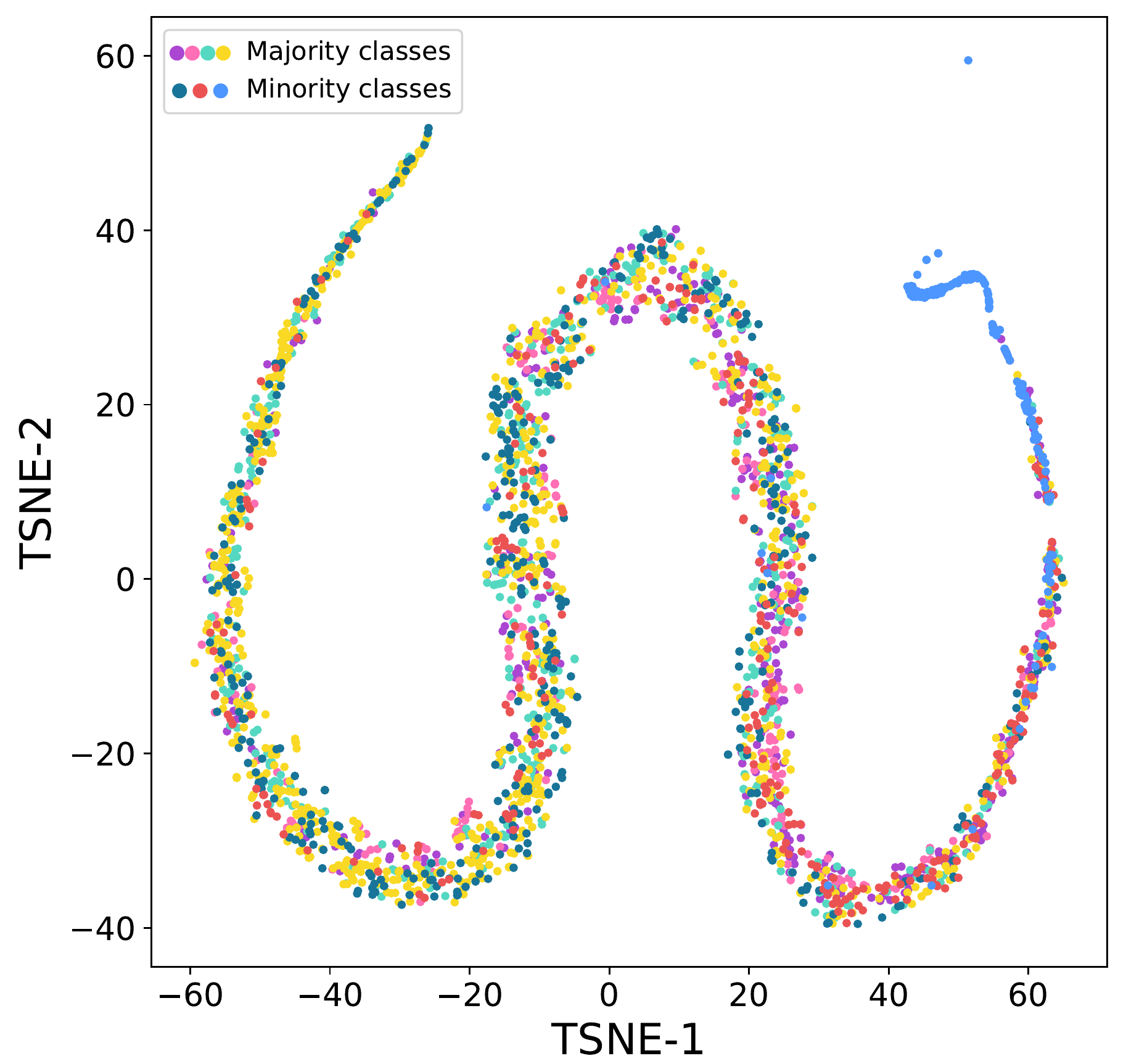}
  }
  \subfigure[GraphENS]{
  \includegraphics[scale=0.2]{visual_ens.pdf}
  }\hspace{-2mm}
  \subfigure[GraphSANN]{
  \includegraphics[scale=0.2]{visual_ours.pdf}
  }
  \caption{Visualization of latent embeddings learned by vanilla GCN, ImGAGN, GraphENS, and GraphSANN.}
  \label{fig:vis-4}
\end{figure}

\xhdr{Class-wise performance Comparison} To further evaluate the performance of our model on classifying nodes from minority classes, we provide the test accuracy curves for all classes and for minority classes only, respectively. Figure \ref{fig:test_curve}(a) shows the accuracy curve including all classes and Figure\ref{fig:test_curve}(b) shows the curve including only minority classes. Here we use Cora as the dataset which has overall 7 node classes, including 3 minority classes. It can be observed that although \name only slightly outperforms the most competitive baseline \textit{GraphENS} in terms of total class accuracy, \name has much better classification results for the three minority classes. The learning curve of \name is also smoother with less vibration in the early process of convergence compared to \textit{GraphENS}.

\xhdr{Visualization} We also compare two extra baselines \textit{ImGAGN} and \textit{Vanilla GCN} for embedding visualization task besides \textit{GraphENS} and \name. The visualization results of all four models are shown in Figure \ref{fig:vis-4}. We can observe that due to lacking additional strategies to handle the class-imbalance problem, the embeddings learned by \textit{Vanilla GCN} can separate some majority classes well (e.g. pink and cyan dots) but mix the clusters of minority classes with other classes. \textit{ImGAGN} surprisingly acquires the worst visualization results. Except for the blue dots, all the other classes are mixed together. We assume it is because \textit{ImGAGN} is originally designed for binary node classification and cannot learn embeddings that separate all the classes well if directly extended to multi-class node classification.



\subsection{Algorithm}\label{sub:algo}
In this subsection, we illustrate the forward propagation procedure of \name in Algorithm \ref{alg:1}.
\SetKwInput{KwInit}{Initialization}
\begin{algorithm*}[h]
\caption{Forward propagation of \name }\label{alg:1}
  \KwIn{The imbalanced graph $\mathcal{G}=\{\mathcal{V},\mathcal{E}, \mathbf{X}\}$;\\
  }
  \KwOut{Predicted node labels.\\}
  \KwInit{Randomly initialize the parameters of unified feature mixer, multi-filter subgraph encoder, adaptive subgraph extractor and multi-filter node classifier;$S_{pair}=\emptyset$; $\mathcal{V}_{syn}=\emptyset$; $\mathcal{E}_{syn}=\emptyset$; \\}
  \For{ m in minority class set $\mathcal{C}_M$}{
  \For{s in $|\mathcal{V}_m|\cdot\zeta$}{
  \tcc{unified node pair sampling}
  $v_s\sim p(\mu\mid\mathcal{C}_M)=\frac{1}{|\mathcal{V}_m|}$;\\
  $v_t\sim p(\mu\mid\mathcal{C})=\frac{\text{log}(|\mathcal{V}_c| + 1)}{(|\mathcal{V}_c|+1)\sum_{c\in\mathcal{C}}\text{log}(|\mathcal{V}_c|+1)}$;\\
  $S_{pair}\leftarrow S_{pair}\cup\{<v_s, v_t>\mid v_s\in\mathcal{V}_m,v_t\in\mathcal{V}\}$;\\
  }
  }
  \For{$v_s, v_t\in S_{pair}$}{
  \tcc{integrated gradient-based feature mixup}
  \textup{Compute integrated gradient along every dimension of the input node features:} $\text{IG}_i(\mathbf{x}) \leftarrow \mathbf{x}_i\int^1_{t=0}\frac{\partial \mathcal{L}_{cls}(t\mathbf{x},\mathbf{y})}{\partial\mathbf{x}_i}dt$;\\
  $\mathbf{D}_t\leftarrow[\text{IG}_1(\mathbf{x}_t),\ldots,\text{IG}_d(\mathbf{x}_t)]$;\\
  \textup{Compute feature similarities:} $\hat{\psi}_{st} \leftarrow \frac{1}{1+\psi_{st}}$; $\psi_{st} \leftarrow \large\Vert \mathbf{W}_p\mathbf{x}_s-\mathbf{W}_p\mathbf{x}_t \large\Vert_{2}$;\\
  \textup{Compute feature mask} $\mathbf{M}_t \leftarrow 1_{\mathbb{R}^+}(\kappa\hat{\psi_{st}}\cdot\mathbf{I}_t-\mathbf{D}_t))$;\\
  \textup{Generate synthetic node feature} $\mathbf{x}_{syn} \leftarrow (1-\mathbf{M}_t)\odot \mathbf{x}_s + \mathbf{M}_t\odot\mathbf{x}_t.$; $\mathcal{V}_{syn}\leftarrow\mathcal{V}_{syn}\cup v_{syn}$\\
  \textup{Sample candidate edges}: $\mathcal{E}_{syn} \leftarrow \mathcal{E}_{syn} \cup \{(v_{syn},u)\mid u\in\mathcal{V}_{nei}\}; \mathcal{V}_{nei} \leftarrow\left[\mathcal{N}_{1}(v_s)\cup \mathcal{N}_{1}(v_t)\right]_{\xi}$;\\
  }
  \For{$e=(v,u)\in\mathcal{E}_{syn}$}{
  \tcc{adaptive subgraph extractor}
  \For{$k\in\mathcal{N}_h(v)\cup\mathcal{N}_h(u)$}{
  \textup{Compute relevance score $f_{rel}(k)$ for every neighbor k};\\
  }
  \textup{Select the top M nodes w.r.t. $f_{rel}(k)$ along with v and u as the subgraph $\mathcal{G}_e=\{\mathcal{V}_e,\mathcal{E}_e\}$};\\
  \For{$l=0,1,\ldots,L$}{
  \For{$u\in\mathcal{V}_e$}{
  \tcc{multi-filter subgraph encoder}
  \textup{Compute weight vector w.r.t. different frequencies:} $\boldsymbol{\alpha}^{(l)}_{(u,k)} \leftarrow \textup{Softmax}(\big[\alpha^{(l)}_{L,(u,k)}, \alpha^{(l)}_{H,(u,k)}, \alpha^{(l)}_{I,(u,k)}\big])$;\\
  $\alpha^{(l)}_{L,(u,k)} \leftarrow \sigma\Big(\mathbf{g}^\mathrm{T}_L\big[\mathbf{W}^{(l)}_L\mathbf{h}^{(l-1)}_u\parallel\mathbf{W}^{(l)}_L\mathbf{h}^{(l-1)}_k\big]\Big)$;\\
     $\alpha^{(l)}_{H,(u,k)} \leftarrow \sigma\Big(\mathbf{g}^\mathrm{T}_H\big[-\mathbf{W}^{(l)}_H\mathbf{h}^{(l-1)}_k\big]\Big)$; \\
     $\alpha^{(l)}_{I,(u,k)} \leftarrow \sigma\Big(\mathbf{g}^\mathrm{T}_I\big[\mathbf{W}^{(l)}_I\mathbf{h}^{(l-1)}_u\big]\Big)$;\\
  \textup{Fuse different frequencies of messages into target node embedding:}\\
  $\mathbf{h}^{(l)}_u \leftarrow \omega\mathbf{h}^{(l-1)}_u+\sum_{k\in \mathcal{N}_1(u)}\boldsymbol{\alpha}^{(l)}_{(u,k)}\mathbf{H}^{(l)}_k$;\\
    $\mathbf{H}^{(l)}_k \leftarrow \textup{ReLU}\Big(\big[\mathbf{W}^{(l)}_L\mathbf{h}^{(l-1)}_k,\mathbf{W}^{(l)}_H\mathbf{h}^{(l-1)}_k,\mathbf{W}^{(l)}_I\mathbf{h}^{(l-1)}_k\big]^{\mathrm{T}}\Big)$;\\
  }
  }
  \textup{Mean readout to compute edge existence probability:}\\
  $\mathbf{h}_\mu \leftarrow \mathbf{h}_\mu^{(1)}\parallel\mathbf{h}_\mu^{(2)}\parallel\ldots\parallel\mathbf{h}_\mu^{(L)}$;\\
   $p_e \leftarrow \frac{1}{|\mathcal{V}_e|}\sum_{u\in\mathcal{V}_e}\mathbf{W}_{pool}\mathbf{h}_u$;\\
  }
\textup{Filter out edges $e\in\mathcal{E}_{syn}$ whose $p_e$ is lower than threshold $\eta$};\\
  $\mathcal{V}_{bal} \leftarrow \mathcal{V} \cup\mathcal{V}_{syn}$;
  $\mathcal{E}_{bal}\leftarrow\mathcal{E}\cup\mathcal{E}_{syn}$;
  $\mathbf{X}_{bal}\leftarrow \textup{CONCAT}(\mathbf{X},\mathbf{X}_{syn})$;\\
  \textup{Encode the balanced graph $\mathcal{G}_{bal}=\{\mathcal{V}_{bal},\mathcal{E}_{bal},\mathbf{X}_{bal}\}$ with a multi-filter node classifier $f$ to predict node classes:}\\
  $\hat{\mathbf{Y}} = \textup{Softmax}(\textup{MLP}(f(\mathcal{G}_{bal}))).$\\
  \Return{\textup{Predicted labels} $\hat{\mathbf{Y}}$}
\end{algorithm*}

\end{document}